\documentclass{article}
\usepackage{graphicx}
\usepackage{subfig}
\usepackage{booktabs}
\usepackage{multirow}
\usepackage[table]{xcolor}
\usepackage{verbatim}

\definecolor{red_boston}{RGB}{204,0,0}
\definecolor{red_stanford}{RGB}{140,21,21}
\definecolor{red_mit}{RGB}{163,31,52}

\makeatletter
\newcommand{\redrowstyle}[0]{\rowstyle{\color{red_boston}}}
\newcommand*{\@rowstyle}{}
\newcommand*{\rowstyle}[1]{
  \gdef\@rowstyle{#1}%
  \@rowstyle\ignorespaces%
}

\newcolumntype{=}{
  >{\gdef\@rowstyle{}}%
}

\newcolumntype{+}{
  >{\@rowstyle}%
}
\title{The UEA multivariate time series classification archive, 2018 }
\author{Anthony Bagnall \and Hoang Anh Dau \and Jason Lines \and Michael Flynn \and James Large \and Aaron Bostrom \and Paul Southam \and Eamonn Keogh }
\date{October 2018}

\begin{document}

\maketitle

\section{Introduction}
In 2002, the UCR time series classification archive was first released with sixteen datasets. It gradually expanded, until 2015 when it increased in size from 45 datasets to 85 datasets. In October 2018 more datasets were added, bringing the total to 128~\cite{dau18archive}. The new archive contains a wide range of problems, including variable length series, but it still only contains univariate time series classification problems. One of the motivations for introducing the archive was to encourage researchers to perform a more rigorous evaluation of newly proposed time series classification (TSC) algorithms. It has worked: most recent research into TSC uses all 85 datasets to evaluate algorithmic advances~\cite{bagnall17bakeoff}. Research into multivariate time series classification, where more than one series are associated with each class label, is in a position where univariate TSC research was a decade ago. Algorithms are evaluated using very few datasets and claims of improvement are not based on statistical comparisons. Recent research has improved somewhat because of the assembly of an archive of 12 datasets by Mustafa Baydogan\footnote{http://www.mustafabaydogan.com/multivariate-time-series-discretization-for-classification.html}. This archive is useful,  but it has limitations. The data, summarised in Table~\ref{tab:baydogan}, are all very small, are not independent and are not representative of many important multivariate time series classification (MTSC) domains.
\begin{table}
        \caption{Datasets that form the current best publicly available collection of multivariate time series classification problems, taken from www.mustafabaydogan.com}
    \label{tab:baydogan}
    \scalebox{0.7}{
\begin{tabular}{ =l +c +c +c +c +c +c | +>{\bf}c}
        & \#of & \# of &  & \multicolumn{2}{c}{Dataset Size} &  &  \\
        & classes & variables & Length & Train & Test & CV & \textnormal{Source} \\
    \midrule
    AUSLAN                  & 95    & 22    & 45-136    &   1140    & 1425  & 10-fold   & \multirow{10}{*}{UCI} \\
    Pendigits               & 10    & 2     & 8         & 300       &   10692   &   &   \\
    Japanese Vowels         & 9     & 12    & 7-29  & 270    & 370  &       & \\
    \cmidrule(r){1-7}
    Robot Failure           &       &       &           &           &       &           &       \\
    \multicolumn{1}{c}{LP1} & 4     & 6     & 15        &   38      & 50    & 5-fold    &    \\
    \multicolumn{1}{c}{LP2} & 5     & 6     & 15        &   17      & 30    &           &       \\
    \multicolumn{1}{c}{LP3} & 4     & 6     & 15        &   17      & 30    &           &       \\
    \multicolumn{1}{c}{LP4} & 3     & 6     & 15        &   42      & 75    &           &       \\
    \multicolumn{1}{c}{LP5} & 5     & 6     & 15        &   64      & 100   &           &       \\
    \cmidrule(r){1-6}
    ECG                     & 2     & 2     & 39-152    &   100     & 100   & 10-fold   & \multirow{2}{*}{Olszewski}    \\
    Wafer                   & 2     & 6     & 104-198   &   298     & 896   &           &       \\
    \cmidrule(r){1-6}
    CMU\_MOCAP\_S16         & 2     & 62    & 127-580   &   29      & 29    & 10-fold   & CMU MOCAP  \\
    \midrule
    ArabicDigits            & 10    & 13    & 4-93      & 6600    & 2200  & x         & \multirow{3}{*}{UCI}    \\
    CharacterTrajectories   & 20    & 3     & 109-205   & 300     & 2558  & x         &    \\
    LIBRAS                  & 15    & 2     & 45        & 180     & 180   & x         &       \\
    \cmidrule(r){1-7}
    uWaveGestureLibrary     & 8     & 3     & 315       & 200     & 4278  & x         & UCR   \\
    \midrule
    \redrowstyle
    PEMS                    & 7     & 963   & 144       & 267     & 173   & x         & UCI   \\
    \redrowstyle
    KickvsPunch             & 2     & 62    & 274-841   & 16      & 10    & x         & CMU MOCAP   \\
    \redrowstyle
    WalkvsRun               & 2     & 62    & 128-1918  & 28      & 16    & x         &    \\
    \redrowstyle
    Network Flow            & 2     & 4     & 50-997    & 803     & 534   & x         & Subakan et al.   \\
    \redrowstyle
    DigitsShape             & 4     & 2     & 30-98     & 24      & 16    & x         &    \\
    \redrowstyle
    Shapes                  & 3     & 2     & 52-98     & 18      & 12    & x         &\\
\end{tabular}}
\end{table}

We aim to address this problem by forming the first iteration of the MTSC archive, to be hosted at the website www.timeseriesclassification.com. Like the univariate archive, this formulation was a collaborative effort between researchers at the University of East Anglia (UEA) and the University of California, Riverside (UCR). The 2018 vintage consists of 30 datasets with a wide range of cases, dimensions and series lengths. For this first iteration of the archive we format all data to be of equal length, include no series with missing data and provide train/test splits. Some of these are also in the Baydogan archive, but the majority have never been used in the context of time series classification before.

The data characteristics are presented in Table~\ref{tab:datasummary}. The whole archive is available as a single zip file~\footnote{www.timeseriesclassification.com/Downloads/MultivariateTSCProblems.zip} (it is over 2GB). The download includes a directory for each problem. In that directory are text files in Weka multi-instance format. We have also provided files for each dimension separately, except for the very high dimensional files where creating thousands of extra files would massively increase the size of the overall archive. Individual problems can be downloaded from the website and code to split multivariate ARFF is available in the codebase.

\begin{table}[!htb]
    \centering
    \resizebox{\linewidth}{!}{
        \begin{tabular}{l|ccccc}
            \hline
            Dataset     &  Train Cases & Test Cases & Dimensions & Length & Classes\\
            \hline
            ArticularyWordRecognition   &   275 &   300 &       9       &   144             &   25\\
            AtrialFibrillation           &   15              &   15              &       2       &   640             &   3\\
            BasicMotions                &   40              &   40              &       6       &   100             &   4\\
            CharacterTrajectories       &   1422            &   1436            &       3       &   182             &   20\\
            Cricket                     &   108             &   72              &       6       &   1197            &   12\\
            DuckDuckGeese               &   60              &   40              &       1345    &   270             &   5\\
            EigenWorms                  &   128             &   131             &       6       &   17984           &   5\\
            Epilepsy                    &   137             &   138             &       3       &   206             &   4\\
            EthanolConcentration   &   261   &   263   &   3   &   1751   &   4\\
            ERing   &   30   &   30   &   4   &   65   &   6\\
            FaceDetection   &   5890   &   3524   &   144   &   62   &   2\\
            FingerMovements   &   316   &   100   &   28   &   50   &   2\\
            HandMovementDirection   &   320   &   147   &   10   &   400   &   4\\
            Handwriting   &   150   &   850   &   3   &   152   &   26\\
            Heartbeat   &   204   &   205   &   61   &   405   &   2\\
            JapaneseVowels   &   270   &   370   &   12   &   29   &   9\\
            Libras   &   180   &   180   &   2   &   45   &   15\\
            LSST   &   2459   &   2466   &   6   &   36   &   14\\
            InsectWingbeat   &   30000   &   20000   &   200   &   78   &   10\\
            MotorImagery   &   278   &   100   &   64   &   3000   &   2\\
            NATOPS   &   180   &   180   &   24   &   51   &   6\\
            PenDigits   &   7494   &   3498   &   2   &   8   &   10\\
            PEMS-SF   &   267   &   173   &   963   &   144   &   7\\
            Phoneme   &   3315   &   3353   &   11   &   217   &   39\\
            RacketSports   &   151   &   152   &   6   &   30   &   4\\
            SelfRegulationSCP1   &   268   &   293   &   6   &   896   &   2\\
            SelfRegulationSCP2   &   200   &   180   &   7   &   1152   &   2\\
            SpokenArabicDigits   &   6599   &   2199   &   13   &   93   &   10\\
            StandWalkJump   &   12   &   15   &   4   &   2500   &   3\\
            UWaveGestureLibrary   &   120   &   320   &   3   &   315   &   8\\
            \hline
            \end{tabular}
    } 
    \caption{A summary of the 30 datasets in the UEA Multivariate Time Series Classification archive, 2018 }
    \label{tab:datasummary}
\end{table}

Weka multi-instance format works well for MTSC when all the series are the same length. It involves defining a relational attribute, which can have multiple occurrences, each separated by a new line marker. So, for example a data file may begin as follows.

\begin{verbatim}
@relation input
@attribute input relational
    @attribute t1 numeric
    @attribute t2 numeric
    @attribute t3 numeric
    @attribute t4 numeric
    @attribute t5 numeric
    @attribute t6 numeric
    @attribute t7 numeric
    @attribute t8 numeric
@end input
@attribute class {0,1,2,3,4,5,6,7,8,9}
@data
"47,27,57,26,0,56,100,40\n100,81,37,0,23,53,90,98",8
\end{verbatim}
This header defines that each series is of length 8, and the number of series per case is defined by the data as two (because there is a single newline). It is a little confusing in code, because each \texttt{Instance} object (i.e. case) contains an \texttt{Instances} object for the relational attribute. For example,

\begin{verbatim}
Instances train= //All the instances
Instance first=train.instance(0); //Get first instance
Instances x= first.relationalValue(0);//Get relational data
Instance s1=x.instance(0);//First series
Instance s2=x.instance(1);//Second series

\end{verbatim}
Example code to manipulate instances is available in the repository\footnote{https://bitbucket.org/TonyBagnall/time-series-classification}. We have done the minimum pre-processing possible, and if the dataset donators provided a train/test split, we have retained that. The sources for these data are numerous and include: the UCI Machine Learning archive; a series of Brain Computer Interface competitions; Kaggle competitions; and some made by us. We split the problems into groups based on the area of application: Human Activity Recognition (HAR) is the largest group (9 problems); Motion classification (4 problems); ECG classification (3 problems); EEG/MEG classification (6 problems); Audio Spectra Classification (5 problems); and others (3 problems).

\newpage
\section{Human Activity Recognition}
Human Activity Recognition (HAR) is the problem of predicting an activity (the class value) based on accelerometer and/or gyroscope data. The data are either three or six dimensions of co-ordinates. HAR is a very popular research area and it is easy to obtain or generate data from this domain. We have included 9 HAR problems. We could have included many more, but we do not want to formulate an archive of just HAR problems until we have enough data from other domains to balance.

\subsection{BasicMotions}

The data was generated as part of a student project in 2016 where four students performed four activities whilst wearing a smart watch.
The watch collects 3D accelerometer and a 3D gyroscope data. It consists of four classes, which are standing, walking, running and
playing badminton. Participants were required to record motion a total of five times, and the data is sampled at 10 Hz for a ten second period.

\begin{figure}[htb]
    \centering
        \subfloat{\includegraphics[trim=0cm -1.8cm 0cm 0cm, width=.5\columnwidth]{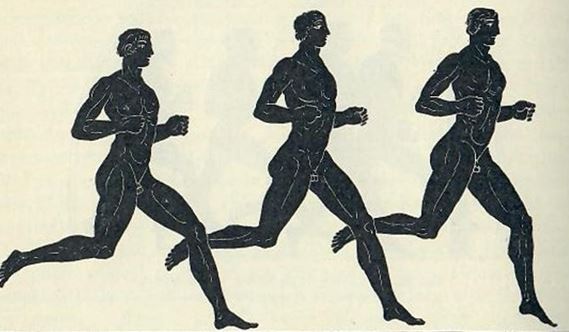}}
        \subfloat{\includegraphics[trim=0cm 0.1cm 0cm 0cm, clip, width=.5\columnwidth]{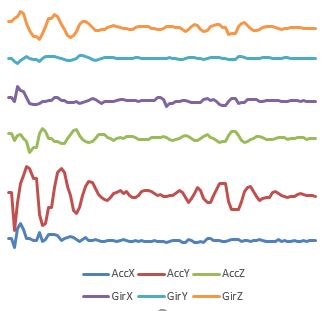}}
    \caption{First train case for the problem BasicMotion. The class label for this case is Standing.}
    \label{fig:BasicMotion}
\end{figure}

\subsection{Cricket}

Cricket requires an umpire to signal different events in the game to a distant scorer. The signals are communicated with motions of the hands. For example, No-Ball is signaled by touching each shoulder with the opposite hand, and TV-Replay, a request for an off-field review of the video of a play, is signaled by miming the outline of a TV screen.

The dataset introduced in Ko {\em et al.} (2005)~\cite{ko2005online} consists of four umpires performing twelve signals, each with ten repetitions. The data, recorded at a frequency of 184 Hz,was collected by placing accelerometers on the wrists of the umpires. Each accelerometer has three synchronous measures for three axes (x, y and z). Thus, we have a
six-dimensional problem from the two accelerometers. Cricket was first formatted for MTSC in~\cite{shokoohi17generalizing}.

\begin{figure}[htb]
    \centering
    \subfloat{\includegraphics[trim=0cm -1.5cm 0cm 0cm, clip, width=.5\columnwidth]{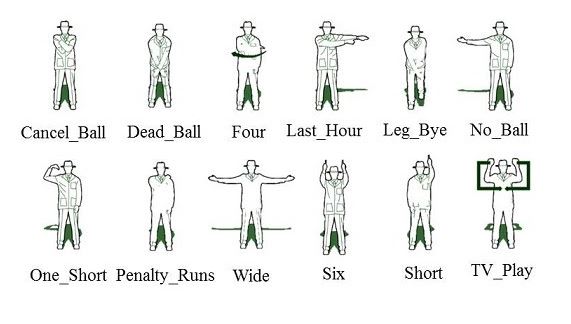}}
    \subfloat{\includegraphics[width=.5\columnwidth]{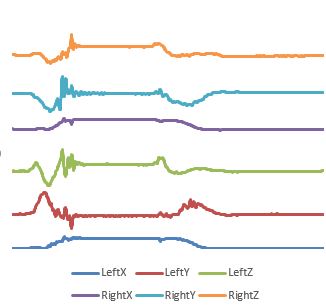}}
    \caption{Image of the class labels and the first train case for the problem Cricket. The class label for this case is Cancel Ball (1).}
    \label{fig:cricket}
\end{figure}

\subsection{Epilepsy}

The data, presented in~\cite{villar2016generalized}, was generated with healthy participants simulating the
class activities. Data was collected from 6
participants using a tri-axial accelerometer on the dominant wrist
whilst conducting 4 different activities. The four tasks, each of
different length, are: WALKING includes different paces and
gestures: walking slowing while gesturing, walking slowly, walking
normal and walking fast, each of 30 seconds long; RUNNING includes
running a 40 meters long corridor; SAWING with a saw and during 30
seconds; and SEIZURE MIMICKING whilst seated, with 5-6 sec before and 30 sec
after the mimicked seizure. The seizure was 30 sec long. Each participant performs each activity 10 times at least. The
mimicked seizures were trained and controlled, following a protocol
defined by an medical expert. All the activities were carried out
indoors, either inside an office or in the corridor around it.

The sampling frequency was 16 Hz. Some activities lasted about 30
seconds, others are 1 minute long, others are about 2 minutes. Our
standard practice for the archive is to truncate data to the length
of the shortest series retained. We removed prefix and suffix flat
series and truncated to the shortest series (approximately 13 seconds), taking a random interval of activity for series
longer than the minimum. A single case
from the original (ID002 Running 16) was removed because the data
was not collected correctly. After tidying the data we have a total
of 275 cases. The train test split is divided into three
participants for training, three for testing, with the IDs removed
for consistency with the rest of the archive.


\begin{figure}[htb]
    \centering
    \subfloat{\includegraphics[width=.5\columnwidth]{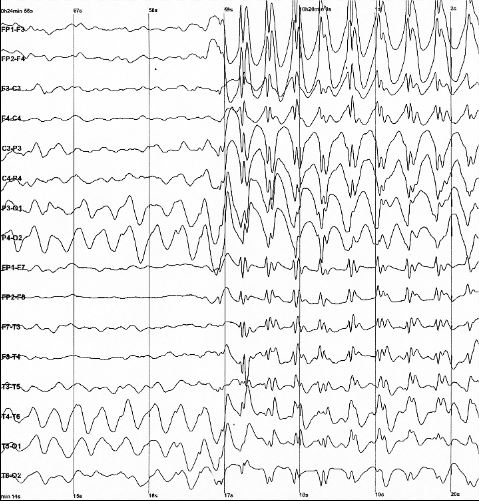}}
    \subfloat{\includegraphics[width=.5\columnwidth]{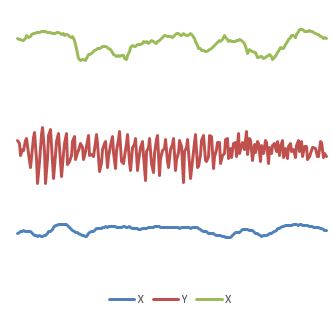}}
    \caption{Example of an Epilepsy EEG and the first train case for the HAR problem Epilepsy. The class label for this case is Epilepsy.}
    \label{fig:epilepsy}
\end{figure}

\subsection{ERing}

This data is generated with a prototype finger ring, called eRing~\cite{wilhelm2015ering}, that can be used to detect hand and finger gestures. eRing uses electric field sensing. The dataset we used to form the archive set is the D dataset used for Finger Posture Recognition. There are six classes for six postures involving the thumb, the index finger, and the middle finger. The data is four dimensional. Each series contains 65 observations. Each series is a measurement from an electrode which varies dependent on the distance to the hand.


\begin{figure}[!ht]
    \centering
    \subfloat{\includegraphics[width=.5\columnwidth]{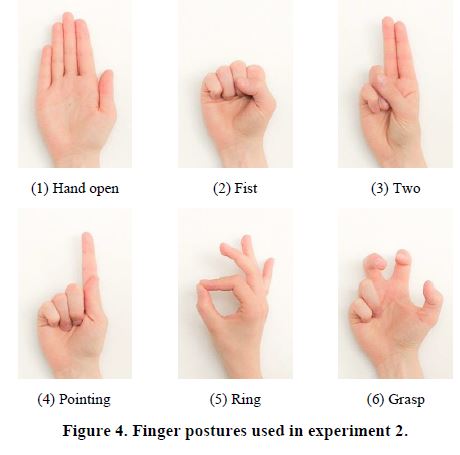}}
    \subfloat{\includegraphics[width=.5\columnwidth]{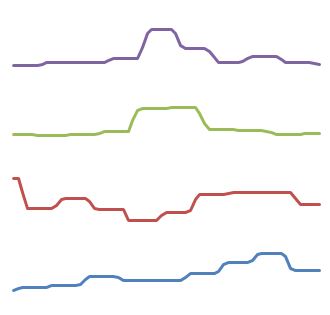}}
    \caption{Image of the E-Ring and the first train case for the HAR problem ERing. The class label for this case is Fist (2).}
    \label{fig:ering}
\end{figure}
\newpage
\subsection{Handwriting}

\begin{figure}[!hb]
    \centering
    \subfloat{\includegraphics[width=.5\columnwidth]{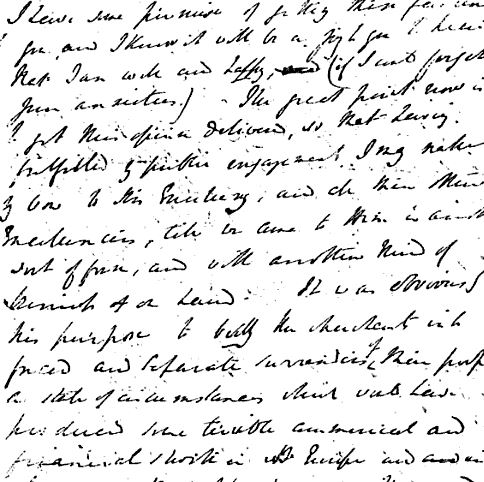}}
    \subfloat{\includegraphics[width=.5\columnwidth]{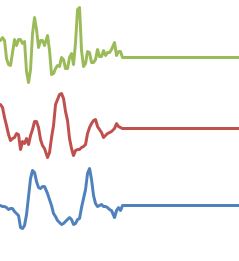}}
    \caption{The first train case for the HAR problem Handwriting. The class label for this case is U (21).}
    \label{fig:Handwriting}
\end{figure}

A dataset of motion taken from a smart watch whilst the subject writes the 26 letters of the alphabet created at UCR and reported in~\cite{shokoohi17generalizing}. There are 150 train cases and 850 test cases. The three dimensions are the three accelerometer values. The data has been padded by those who donated it (see Figure~\ref{fig:Handwriting}).

\subsection{Libras}

The LIBRAS Movement Database is part of the UCI archive and was used in~\cite{dias16algoritmos}. LIBRAS, acronym of the Portuguese name "Lingua BRAsileira de Sinais", is the oficial brazilian sign language. The dataset contains 15 classes of 24 instances each, where each class references to a hand movement type in LIBRAS. The hand movement is represented as a bi-dimensional curve performed by the hand in a period of time. The curves were obtained from videos of hand movements, with the Libras performance from 4 different people, during 2 sessions. Each video corresponds to only one hand movement and has about $7$ seconds.

In the video pre-processing, a time normalization is carried out selecting 45 frames from each video, in according to an uniform distribution. In each frame, the centroid pixels of the segmented objects (the hand) are found, which compose the discrete version of the curve F with 45 points. All curves are normalized in the unitary space. In order to prepare these movements to be analysed by algorithms, we have carried out a mapping operation, that is, each curve F is mapped in a representation with 90 features, with representing the coordinates of movement.

Each instance represents 45 points on a bi-dimensional space, which can be plotted in an ordered way (from 1 through 45 as the X co-ordinate) in order to draw the path of the movement.

\begin{figure}[!htb]
    \centering
    \subfloat{\includegraphics[width=.5\columnwidth]{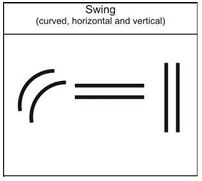}}
    \subfloat{\includegraphics[width=.5\columnwidth]{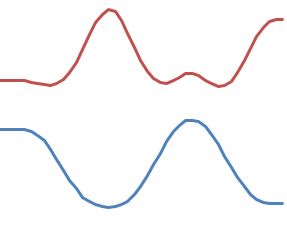}}
    \caption{Example of the first train case for the HAR problem Libras. The class label for this case is 1.}
    \label{fig:Libras}
\end{figure}

\subsection{NATOPS}

This data was originally part of a competition for the AALTD workshop in 2016~\footnote{https://aaltd16.irisa.fr/challenge/} and is described in
~\cite{ghouaiel2017continuous}. The problem is to automatically detect the motion of various Naval Air Training and Operating Procedures Standardization motions used to control plane movements.


\begin{figure}[htb]
    \centering
    \subfloat{\includegraphics[width=.5\columnwidth]{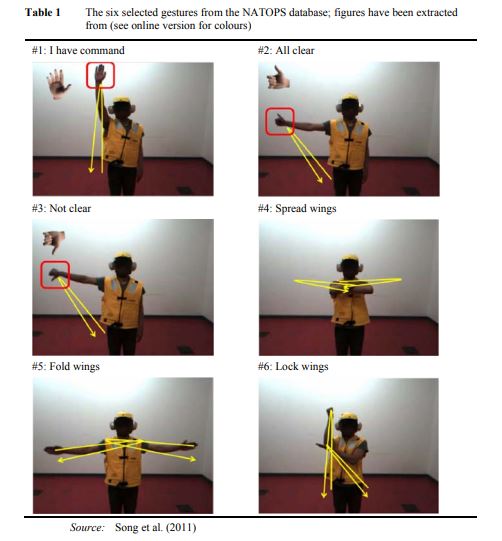}}
    \subfloat{\includegraphics[width=.5\columnwidth]{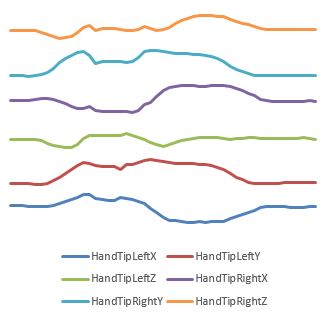}}
    \caption{Examples of the six classes and six series in the first train case for the HAR problem NATOPS. The class label for this case is Spread Wings (4).}
    \label{fig:NATOPS}
\end{figure}

The data is generated by sensors on the hands, elbows, wrists and thumbs. The data are the x, y, z coordinates for each of the eight locations, meaning there are 24 dimensions.  The six classes are separate actions: I have command; All clear; Not clear; Spread wings; Fold wings; and  Lock wings.

\subsection{RacketSports}
The data was created by university students playing badminton or squash whilst wearing a smart watch (Sony Smart watch 3). The watch relayed the x, y, z coordinates for
both the gyroscope and accelerometer to an android phone (One Plus 5). The problem is to identify which sport and which stroke the players are making. The data was collected at a rate of 10 HZ over 3 seconds whilst the player played
either a forehand/backhand in squash or a clear/smash in badminton. The data was collected as part of an undergraduate project by Phillip Perks in 2017/18.

\begin{figure}[!hb]
    \centering
    \subfloat{\includegraphics[width=.5\columnwidth]{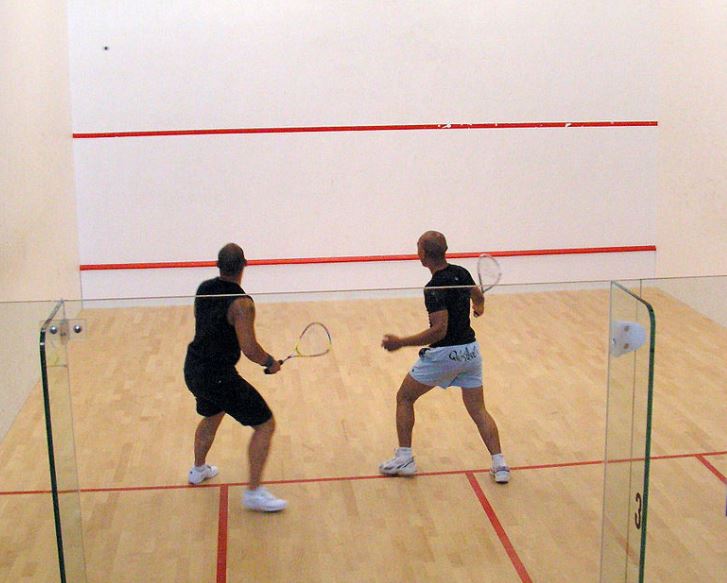}}
    \subfloat{\includegraphics[width=.5\columnwidth]{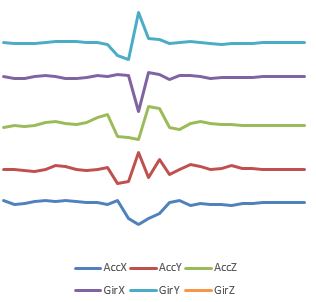}}
    \caption{Example of the first train case for the HAR problem RacketSports. The class label for this case is Badminton Smash.}
    \label{fig:racketsports}
\end{figure}
\newpage
\subsection{UWaveGestureLibrary}

A set of eight simple gestures generated from accelerometers. The data consists of the x, y, z coordinates of each motion. Each series is 315 long. The data was first described in~\cite{liu2009uwave}.


\begin{figure}[!htb]
    \centering
    \subfloat{\includegraphics[trim=0cm -1cm 0cm 0cm, clip, width=.5\columnwidth]{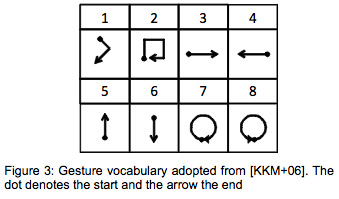}}
    \subfloat{\includegraphics[width=.5\columnwidth]{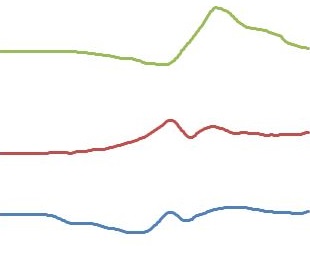}}
    \caption{Example of the eight classes and the first train case for the HAR problem UWaveGestureLibrary. The class label for this case is 1.}
    \label{fig:UWaveGestureLibrary}
\end{figure}

\newpage
\newpage
\section{Motion Classification}

We differentiate HAR data, which characterised by motion recorded by accelerometer and/or gyroscopes, with data recording other forms of movement.

\subsection{ArticularyWordRecognition}

An Electromagnetic Articulograph (EMA) is an apparatus used to measure the movement of the tongue and lips during speech. The motion tracking using EMA is registered by attaching small sensors on the surface of the articulators (e.g., tongue and lips). The spatial accuracy of motion tracking using EMA AG500 is 0.5 mm. This is the EMA dataset used in~\cite{wang2013word}] which contains data collected from multiple native English native speakers producing 25 words. Twelve sensors were used in data collection, each providing x, y and z time-series positions with a
sampling rate of 200 Hz. The sensors are located on the forehead,
tongue; from tip to back in the midline, lips and jaw. The three head sensors (Head Center, Head Right, and Head Left) attached on a pair of glasses were used to calculate head-independent movement of other sensors. Tongue sensors were named T1, T2, T3, and T4, from tip to back. Of the total of 36 available dimensions, this dataset includes just 9, since that was the format of the  data obtained from the  Shokoohi-Yekta {\em et al.}~\cite{shokoohi17generalizing}.

\begin{figure}[!htb]
    \centering
    \subfloat{\includegraphics[trim=0cm -2cm 0cm 0cm, clip, width=.5\columnwidth]{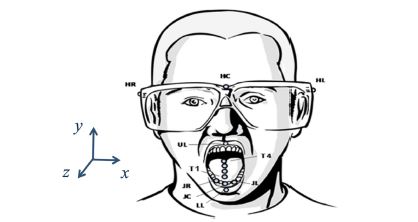}}
    \subfloat{\includegraphics[width=.5\columnwidth]{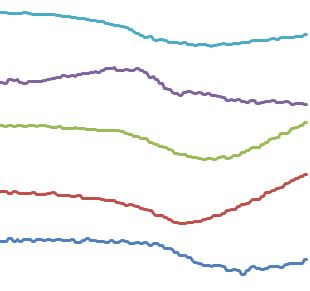}}
    \caption{Example of the first train case for the Motion problem ArticularyWordRecognition. The class label for this case is 1.0.}
    \label{fig:ArticularyWordRecognition}
\end{figure}

\subsection{CharacterTrajectories}

The data were taken from the UCI dataset, provided by Ben Williams, School of Informatics, University of Edinburgh. The data consists of 2858 character samples, captured using a WACOM tablet. Three dimensions were kept - x, y, and pen tip force. The data has been numerically differentiated and Gaussian smoothed,  with a sigma value of 2. Data was captured at 200Hz. The data was normalised. Only characters with a single 'PEN-DOWN' segment were considered. Character segmentation was performed using a pen tip force cut-off point. The characters have also been shifted so that their velocity profiles best
match the mean of the set. The characters here were used for a PhD study on primitive extraction using HMM based models~\cite{williams2008modelling}.

Each instance is a 3-dimensional pen tip velocity trajectory. The original data has different length cases. The class label is one of 20 characters: a; b; c; d; e; g; h; l; m; n; o; p; q; r; s; u; v; w; y; z. To conform with the repository, we have truncated all series to the length of the shortest, which is 182, which will no doubt make classification harder.

\begin{figure}[!htb]
    \centering
    \subfloat{\includegraphics[width=.25\columnwidth]{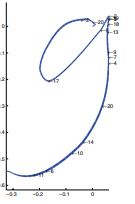}}
    \subfloat{\includegraphics[width=.5\columnwidth]{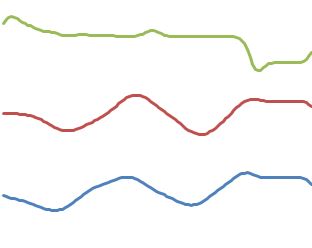}}
    \caption{Example of the first train case for the Motion problem CharacterTrajectories. The class label for this case is g.}
    \label{fig:CharacterTrajectories}
\end{figure}

\subsection{EigenWorms}
Caenorhabditis elegans is a roundworm commonly used as a model organism in the study of genetics. The movement of these worms is known to be a useful
indicator for understanding behavioural genetics. Brown {\em et al.}~\cite{brown2013dictionary} describe a system for recording the motion of worms on an agar plate and measuring a range of human-defined features~\cite{yemini2013database}.  It has been shown that the space of shapes Caenorhabditis elegans adopts on an agar plate can be
represented by combinations of six base shapes, or eigenworms. Once the worm outline is extracted, each frame of worm motion can be captured by six
scalars representing the amplitudes along each dimension when the shape is projected onto the six eigenworms. Using data collected for the work described in~\cite{yemini2013database},  we address the problem of classifying individual worms as wild-type or mutant based on the time series. The data were extracted from the C. elegans behavioural database\footnote{http://movement.openworm.org/}. We have 259 cases, which we split into 131 train and 128 test cases. We have truncated each series to the shortest series, after which each series has 17984 observations. Each worm is classified as either wild-type (the N2 reference strain) or one of four mutant types:
goa-1; unc-1; unc-38 and unc-63.

\begin{figure}[!htb]
    \centering
    \subfloat{\includegraphics[width=.5\columnwidth]{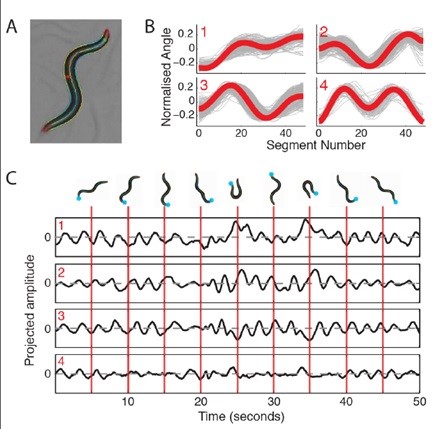}}
    \subfloat{\includegraphics[width=.5\columnwidth]{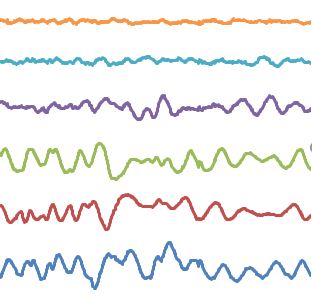}}
    \caption{Example of the first train case for the Motion problem EigenWorms. The class label for this case is wild-type (1).}
    \label{fig:EigenWorms}
\end{figure}

\subsection{PenDigits}

\begin{figure}[!htb]
    \centering
    \subfloat{\includegraphics[trim=0cm -7cm 0cm 0cm, clip, width=.5\columnwidth]{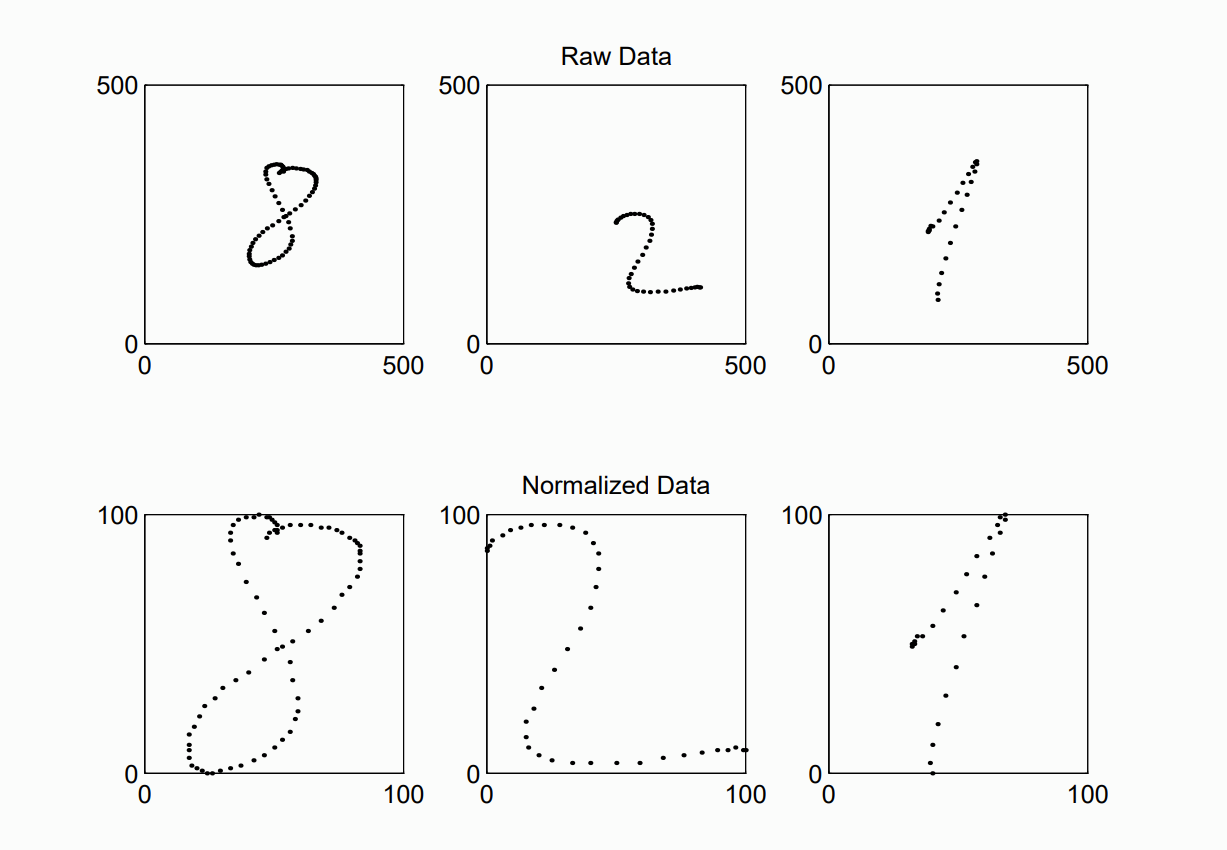}}
    \subfloat{\includegraphics[width=.5\columnwidth]{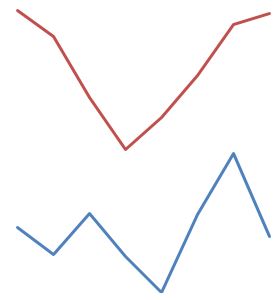}}
    \caption{Example of the first train case for the Motion problem PenDigits. The class label for this case is 8.}
    \label{fig:PenDigits}
\end{figure}

This is a handwritten digit classification task, taken from the UCI Archive \footnote{https://archive.ics.uci.edu/ml/datasets/Pen-Based+Recognition+of+Handwritten+Digits} and originally described in~\cite{alimouglu01combining}. 44 writers were asked to draw the digits 0 to 9, where instances are made up of the x and y coordinates of the pen-tip traced across a digital screen.

The coordinate data were originally recorded at a 500x500 pixel resolution. It was then normalised and sampled to 100x100. Then, based on expert knowledge from the original dataset creators, the data was spatially resampled such that data are sampled with a constant spatial step and variable time step. The data was resampled to 8 spatial points, resulting in each instance having 2 dimensions of 8 points, with a single class label (0\ldots9) being the digit drawn.

\newpage
\section{ECG Classification}
ECG Classification is an obvious application for MTSC. However, we found it surprisingly difficult to find many problems in this domain. The Physionet data often requires bespoke software to process and is not always an obvious classification problem. We hope to get more data in this domain in the future.

\subsection{AtrialFibrillation}
This dataset of two-channel ECG recordings has been created from data used in the Computers in Cardiology Challenge 2004~\footnote{https://www.physionet.org/physiobank/database/aftdb/}, an open competition with the goal of developing automated methods for predicting spontaneous termination of atrial fibrillation (AF).
The raw instances were 5 second segments of atrial fibrillation, containing two ECG signals, each sampled at 128 samples per second. The multivariate data organises these channels such that each is one dimension. The class labels are: n, s and t.
Class n is described as a non termination atrial fibrillation (that is, it did not terminate for at least one hour after the original recording of the data).
class s is described as an atrial fibrillation that self terminates at least one minuet after the recording process. Class t is described as terminating immediately, that is within one second of the recording ending. More details are in~\cite{moody2004spontaneous}.


\begin{figure}[!htb]
    \centering
    \subfloat{\includegraphics[trim=0cm -2.5cm 0cm 0cm, clip, width=.5\columnwidth]{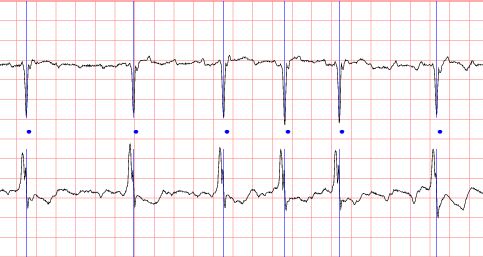}}
    \subfloat{\includegraphics[width=.5\columnwidth]{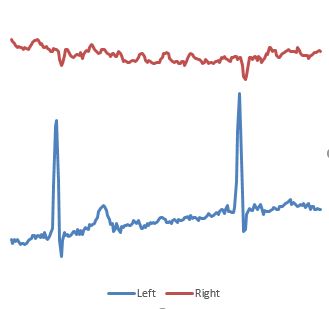}}
    \caption{The first train case for the ECG problem AtrialFibrillation. The class label for this case is `n'.}
    \label{tab:AtrialFibrillation}
\end{figure}

\subsection{StandWalkJump}

This Physionet dataset~\footnote{https://www.physionet.org/physiobank/database/macecgdb/} was presented in~\cite{Behravan2015Rate}. Short duration ECG signals were recorded from a healthy 25-year-old male performing different physical activities to study the effect of motion artifacts on ECG signals and their sparsity. The raw data was sampled at 500 Hz, with a resolution of 16 bits before an analogue gain of 100 and ADC was applied. A spectrogram of each instance was then created with a window size of 0.061 seconds and an overlap of 70\%. Each instance in this multivariate dataset is arranged such that each dimension is a frequency band from the spectrogram. There are three classes, standing, walking and jumping, each consists of 9 instances.


\begin{figure}[!htb]
    \centering
    \subfloat{\includegraphics[trim=0cm -2.5cm 0cm 0cm, clip, width=.5\columnwidth]{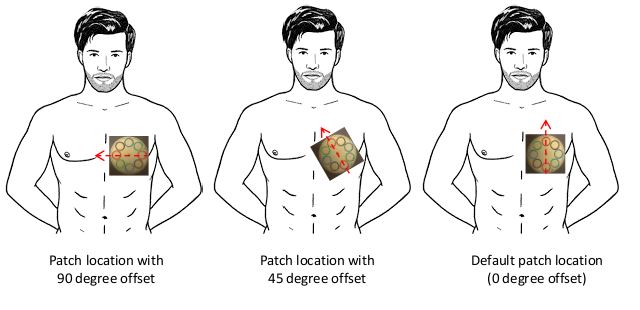}}
    \subfloat{\includegraphics[width=.5\columnwidth]{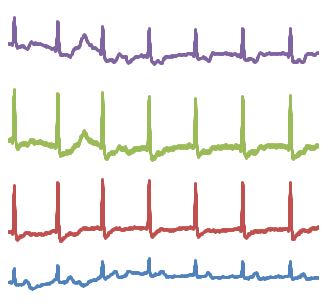}}
    \caption{The first train case for the ECG problem StandWalkJump. The class label for this case is standing.}
    \label{fig:StandWalkJump}
\end{figure}

\newpage
\newpage
\section{EEG/MEG Classification}
Our second largest group of problems, EEG and MEG classification has a wide range of applications in medicine, psychology and human computer interaction. The majority of our data were derived from the Brain Computer Interface competitions\footnote{http://bbci.de/competition}.

\subsection{FingerMovements}

This dataset was provided by Fraunhofer-FIRST, Intelligent Data Analysis Group (Klaus-Robert Müller), and Freie Universität Berlin, Department of Neurology, Neurophysics Group (Gabriel Curio)\footnote{http://www.bbci.de/competition/ii/berlin\_desc.html} and is described in~\cite{blankertz2002classifying}.

This dataset was recorded from a normal subject during a no-feedback session. The subject sat in a normal chair, relaxed arms resting on the table, fingers in the standard typing position at the computer keyboard. The task was to press with the index and little fingers the corresponding keys in a self-chosen order and time, i.e. using self-paced key typing. The experiment consisted of 3 sessions of 6 minutes each. All sessions were conducted on the same day with some minutes break in between. Typing was done at an average speed of 1 key per second.

There are 316 train cases and 100 test cases. Each case is a recording of 28 EEG channels of 500 ms length each ending 130 ms before a key-press.
This is downsampled at 100 Hz (as recommended) so each channel consists of 50 observations. Channels are in the following order:
(F3, F1, Fz, F2, F4, FC5, FC3, FC1, FCz, FC2, FC4, FC6, C5, C3, C1, Cz, C2, C4, C6, CP5, CP3, CP1, CPz, CP2, CP4, CP6, O1, O2).

The recording was made using a NeuroScan amplifier and a Ag/AgCl electrode cap from ECI. 28 EEG channels were measured at positions of the international 10/20-system (F, FC, C, and CP rows and O1, O2). Signals were recorded at 1000 Hz with a band-pass filter between 0.05 and 200 Hz.


\begin{figure}[!htb]
    \centering
    \subfloat{\includegraphics[width=.5\columnwidth]{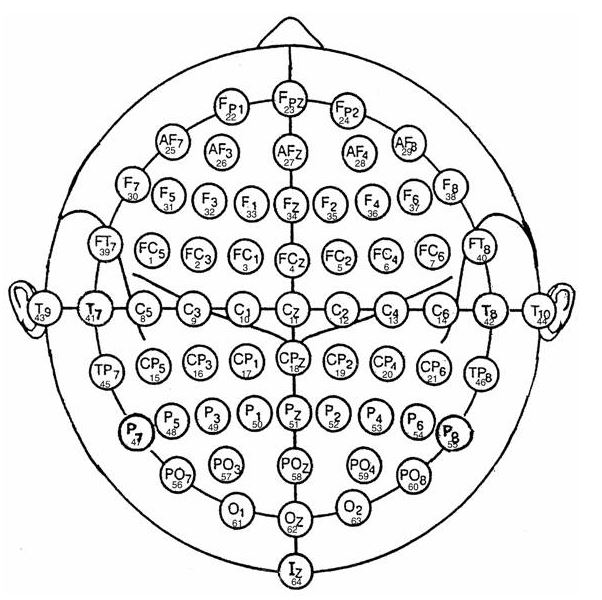}}
    \subfloat{\includegraphics[width=.5\columnwidth]{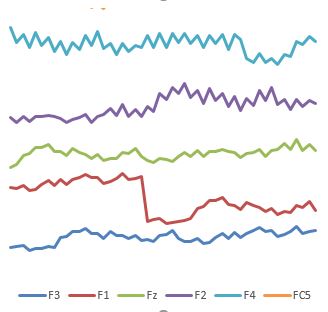}}
    \caption{Example of the first train case for the EEG problem Finger Movements. The class label for this case is `left'.}
    \label{fig:FingerMovements}
\end{figure}

\subsection{MotorImagery}

This is Dataset 1 in BCI III~\footnote{http://bbci.de/competition/iii/desc\_I.html} and is reported in~\cite{lal05methods}, provided by University of Tübingen, Germany, Dept. of Computer Engineering (Prof. Rosenstiel) and Institute of Medical Psychology and Behavioral Neurobiology (Niels Birbaumer), and
Max-Planck-Institute for Biological Cybernetics, Tübingen, Germany (Bernhard Schölkopf), and
Universität Bonn, Germany, Dept. of Epileptology (Prof. Elger). During the BCI experiment, a subject had to perform imagined movements of either the left small finger or the tongue. The time series of the electrical brain activity was picked up during these trials using a 8x8 ECoG platinum electrode grid which
was placed on the contralateral (right) motor cortex. The grid was assumed to cover the right motor cortex completely, but due to its size (approx. 8x8cm), it partly covered also surrounding cortex areas. All recordings were performed with a sampling rate of 1000Hz.
After amplification the recorded potentials were stored as microvolt values. Every trial consisted of either an imagined tongue or an
imagined finger movement and was recorded for 3 seconds duration. To avoid visually evoked potentials being reflected by the data,
the recording intervals started 0.5 seconds after the visual cue had ended. The EEG data has 64 dimensions, each of which is 3000 long (3 seconds measurement).
The train data has 278 cases, the test data 100. The class labels are finger or tongue (the imagined movements). The best submitted solution obtained 91\% accuracy on the test data.

\begin{figure}[!htb]
    \centering
    \subfloat{\includegraphics[width=.5\columnwidth]{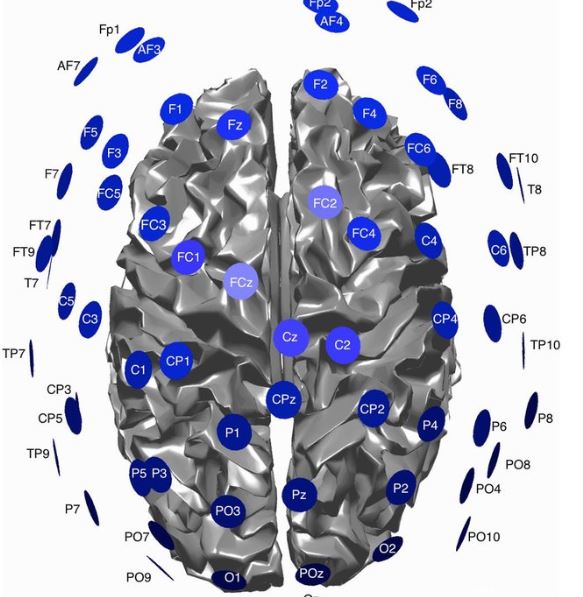}}
    \subfloat{\includegraphics[width=.5\columnwidth]{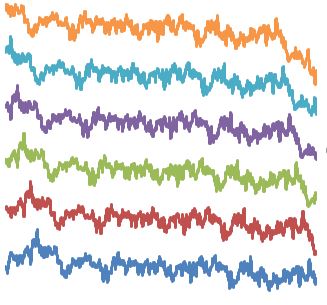}}
    \caption{Example of the first train case for the EEG problem Finger Movements. The class label for this case is `finger'.}
    \label{fig:fingermovements}
\end{figure}

\subsection{SelfRegulationSCP1}

This dataset is Ia in BCI II~\footnote{http://bbci.de/competition/ii/tuebingen\_desc\_i.html} reported in~\cite{birbaumer99spelling}:  Self-regulation of Slow Cortical Potentials.
It was provided by University of Tuebingen. The data were taken from a healthy subject. The subject was asked to move a cursor up and down on a computer screen, while his cortical potentials were taken. During the recording, the subject received visual feedback of his slow cortical potentials (Cz-Mastoids).  Cortical positivity leads to a downward movement of the cursor on the screen.  Cortical negativity leads to an upward movement of the cursor.  Each trial lasted 6s.

During every trial, the task was visually presented by a highlighted goal at either the top or bottom of the screen to indicate negativity or positivity from second 0.5 until the end of the trial. The visual feedback was presented from second 2 to second 5.5. Only this 3.5 second interval of every trial is provided for training and testing. The sampling rate of 256 Hz and the recording length of 3.5s results in 896 samples per channel for every trial.

The train data consists of 268 trials recorded on two different days and mixed randomly. 168 of the overall 268 trials origin from day 1, the remaining 100 trials from day 2.
The data is derived from the two train files Traindata\_0.txt and Traindata\_1.txt.
Each instance has six dimensions (EEG channels above) of length 896. Class labels are negativity or positivity. There are 293 test data, the labels of which were released after the competition. The best approach has an error rate of 11.3\% on the test data (presumably 33 incorrect).

\begin{figure}[!htb]
    \centering
    \subfloat{\includegraphics[width=.5\columnwidth]{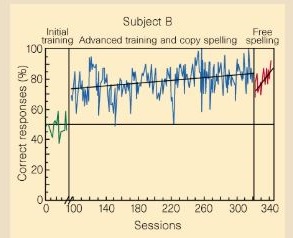}}
    \subfloat{\includegraphics[width=.5\columnwidth]{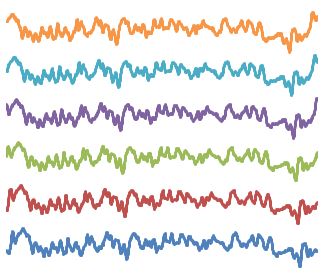}}
    \caption{Example of the first train case for the EEG problem SelfRegulationSCP1. The class label for this case is `negativity'.}
    \label{fig:SelfRegulationSCP1}
\end{figure}

\subsection{SelfRegulationSCP2}

Dataset Ib in BCI II reported in~\cite{birbaumer99spelling}:  Self-regulation of Slow Cortical Potentials.

The datasets were taken from an artificially respirated ALS patient. The subject was asked to move a cursor up and down on a computer screen, while his cortical potentials were taken. During the recording, the subject received auditory and visual feedback of his slow cortical potentials (Cz-Mastoids).  Cortical positivity lead to a downward movement of the cursor on the screen.  Cortical negativity lead to an upward movement of the cursor.  Each trial lasted 8s.
During every trial, the task was visually and auditorily presented by a highlighted goal at the top (for negativity) or bottom (for positivity) of the screen from second 0.5 until second 7.5 of every trial.  In addition, the task ("up" or "down") was vocalised at second 0.5.
The visual feedback was presented from second 2 to second 6.5. Only this 4.5 second interval of every trial is provided for training and testing. The sampling rate of 256 Hz and the recording length of 4.5s results in 1152 samples per channel for every trial.

The train data contains  200 trials, 100 of each class which were recorded on the same day and permuted randomly.
There are 7 dimensions and the series are length 1152.

Test data contains 180 trials of test data.  This test data was recorded after the train data (during the same day) day.  The 180 trials belong to either class 0 or class 1.

Note that it is not clear if there is any information contained in this dataset that is useful for the classification task. A view on the result suggests that it is not. The best has error 45.5\%.

\begin{figure}[!htb]
    \centering
    \subfloat{\includegraphics[width=.5\columnwidth]{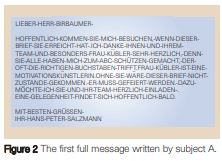}}
    \subfloat{\includegraphics[width=.5\columnwidth]{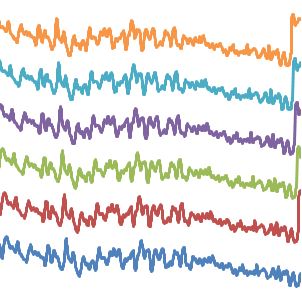}}
    \caption{Example of the first train case for the EEG problem SelfRegulationSCP2. The class label for this case is `negativity'.}
    \label{fig:SelfRegulationSCP2}
\end{figure}

\subsection{FaceDetection}

This data is from the train set of a Kaggle competition\footnote{https://www.kaggle.com/c/decoding-the-human-brain/data}.
It consists of MEG recordings and the class labels (Face/Scramble), from 10 subjects (subject01 to subject10), test data from 6 subjects (subject11 to 16).
For each subject approximately 580-590 trials are available. Each trial consists of 1.5 seconds of MEG recording (starting 0.5sec before the stimulus starts) and the
related class label, Face (class 1) or Scramble (class 0).
The data were down-sampled to 250Hz and high-pass filtered at 1Hz. 306 timeseries were recorded, one for each of the 306 channels, for each trial.
All the pre-processing steps were carried out with mne-python. The trials of each subject are arranged into a 3D data matrix (trial x channel x time) of size 580 x 306 x 375.

\begin{figure}[!ht]
    \centering
    \subfloat{\includegraphics[trim=0cm -2.5cm 0cm 0cm, clip, width=.5\columnwidth]{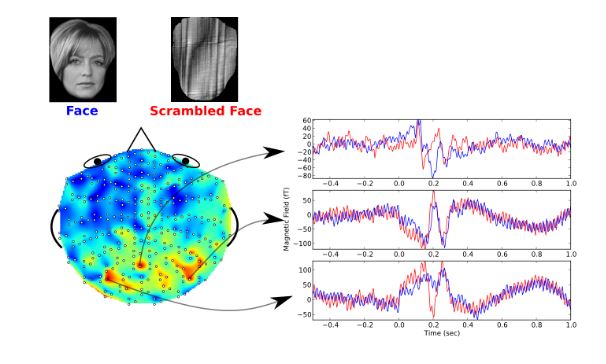}}
    \subfloat{\includegraphics[width=.5\columnwidth]{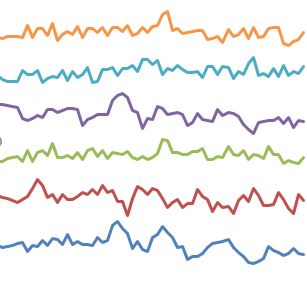}}
    \caption{Example of the first train case for the EEG problem FaceDetection. The class label for this case is 0.}
    \label{fig:EEGFaceDetection}
\end{figure}

\newpage
\subsection{HandMovementDirection}
This is the third dataset from the BCI IV competition\footnote{http://bbci.de/competition/iv/}. It was provided by the Brain Machine Interfacing Initiative, Albert-Ludwigs-University Freiburg, the Bernstein Center for Computational Neuroscience Freiburg and the Institute of Medical Psychology and Behavioral Neurobiology, University of Tübingen (Stephan Waldert, Carsten Mehring, HubertPreissl, Christoph Braun).

Two subjects were recorded moving a joystick with only their hand and wrist in one of four directions (right, up, down, left) of their choice after hearing a prompt. The task is to classify the direction of movement from the Magnetoencephalography (MEG) data recorded during the activity. Each instance contains data from $0.4$s before to $0.6$s after the movement for 10 channels of the MEG reading that are located over the motor areas. Further information about the data collection process can be found at\footnote{http://bbci.de/competition/iv/desc\_3.pdf}.

The train/test split given in this archive corresponds to the exact split provided in the original competition, with the trails for the two subjects merged.

\begin{figure}[!htb]
    \centering
    \subfloat{\includegraphics[trim=0cm -3cm -1cm 0cm, width=.3\columnwidth]{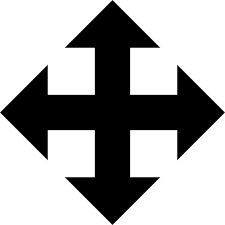}}
    \subfloat{\includegraphics[width=.5\columnwidth]{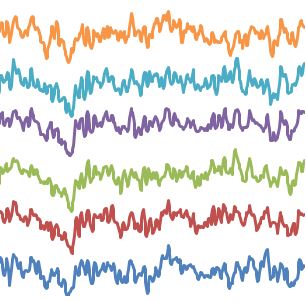}}
    \caption{Example of the first train case for the EEG problem HandMovementDirection. The class label for this case is `1 - right'.}
    \label{fig:HandMovementDirection}
\end{figure}

\newpage
\section{Audio Spectra Classification}

Classification of audio signals is a univariate time series classification problem. However, it is common in this field to run a sliding (or striding) window over the signal and extract the spectra for each window. Each frequency bin then forms a series over the number of windows.

\subsection{DuckDuckGeese}

This dataset was derived from recordings found on the Xeno Canto website\footnote{www.xenocanto.com}. Each recording was taken from either the A or B quality category. Due to the variation in recorded sample rate all recordings were downsampled to 44100Hz using the MATLAB resample function.
Each recording was then center truncated to 5 seconds (length of smallest recording), before being transformed into a spectogram using a window size of 0.061 and an overlap value of 70\%. The classes are as follows:
Black-bellied Whistling Duck (20 instances); Canadian Goose (20 instances);
Greylag Goose (20 instances); Pink Footed Goose (20 instances); and White-faced Whistling Duck (20 instances).

\begin{figure}[!ht]
    \centering
    \subfloat{\includegraphics[trim=0cm -5cm 0cm 0cm, clip, width=.5\columnwidth]{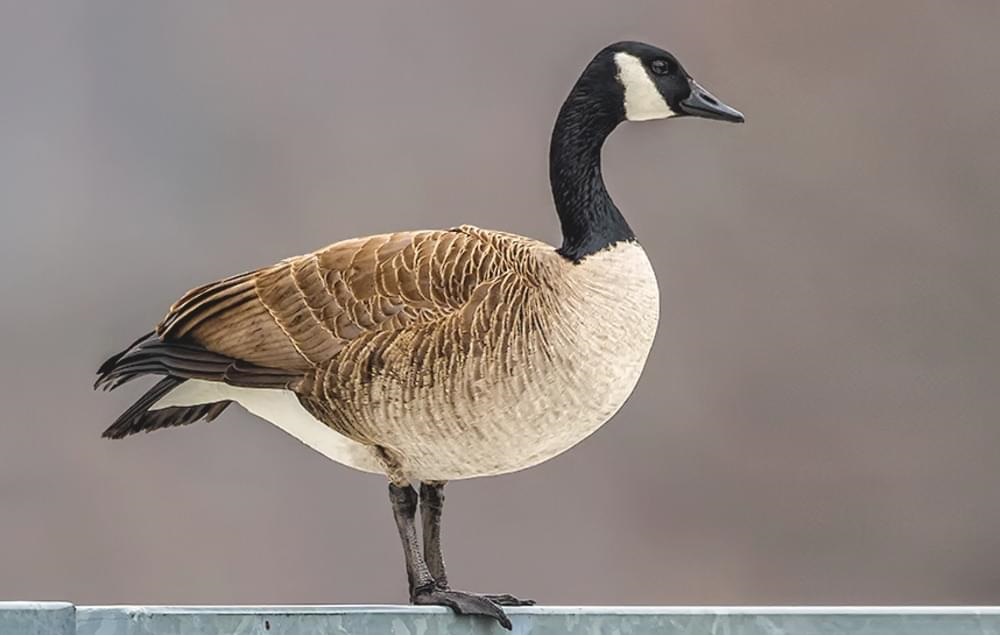}}
    \subfloat{\includegraphics[width=.5\columnwidth]{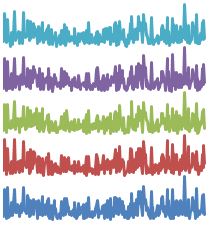}}
    \caption{Example of the first train case for the problem DuckDuckGeese. The class label for this case is Black-bellied Whistling Duck.}
    \label{fig:DuckDuckGeese}
\end{figure}

\newpage
\subsection{Heartbeat}

This dataset is derived from the PhysioNet/CinC Challenge 2016\footnote{https://www.physionet.org/physiobank/database/challenge/2016/}. Heart sound recordings were sourced from several contributors around the world, collected at either a clinical or nonclinical environment, from both healthy subjects and pathological patients. The heart sound recordings were collected from different locations on the body. The typical four locations are aortic area, pulmonic area, tricuspid area and mitral area, but could be one of nine different locations.
The sounds were divided into two classes: normal and abnormal. The normal recordings were from healthy subjects and the abnormal ones were from patients with a confirmed cardiac diagnosis. The patients suffer from a variety of illnesses, but typically they are heart valve defects and coronary artery disease patients. Heart valve defects include mitral valve prolapse, mitral regurgitation, aortic stenosis and valvular surgery. All the recordings from the patients were generally labeled as abnormal. Both healthy subjects and pathological patients include both children and adults.

Each recording was truncated to 5 seconds. A spectrogram of each instance was then created with a window size of 0.061 seconds and an overlap of 70\%.
Each instance in this multivariate dataset is arranged such that each dimension is a frequency band from the spectrogram.
The two classes normal and abnormal consist of 113 and 296 respectively.

\begin{figure}[!htb]
    \centering
    \subfloat{\includegraphics[trim=0cm -1cm 0cm 0cm, clip, width=.5\columnwidth]{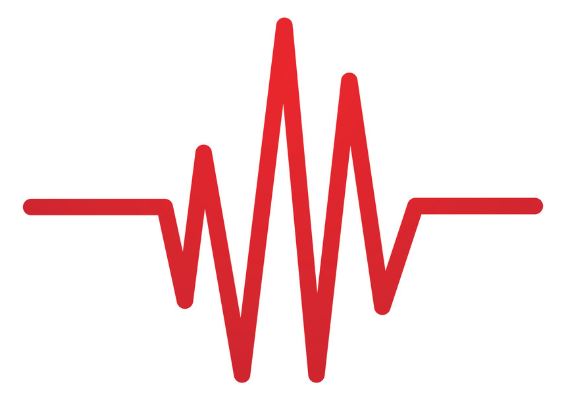}}
    \subfloat{\includegraphics[width=.5\columnwidth]{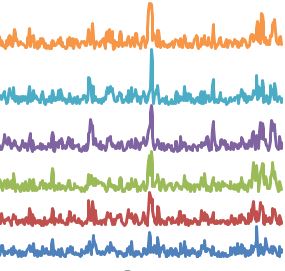}}
    \caption{Example of the first train case for the spectral problem Heartbeat.}
    \label{fig:Heartbeat}
\end{figure}

\subsection{InsectWingbeat}

The InsectWingbeat data was generated by the UCR computational entomology group and used in the paper Flying Insect Classification with Inexpensive Sensors~\cite{chen2014flying}. The original data is a reconstruction of the sound of insects passing through a sensor.
The data in the archive is the power spectrum of the sound.
A spectorgram of each 1 second sound segment was created with a window length of 0.061 seconds and an overlap of 70\%.  Each instance in this multivariate dataset is arranged such that each dimension is a frequency band from the spectrogram. Each of the 10 classes in this dataset consist of 5,000 instances. The 10 classes are male and female mosquitos (Ae. aegypti, Cx. tarsalis, Cx. quinquefasciants, Cx. stigmatosoma), two types of flies (Musca domestica and Drosophila simulans) and other insects.

\begin{figure}[!htb]
    \centering
    \subfloat{\includegraphics[width=.5\columnwidth]{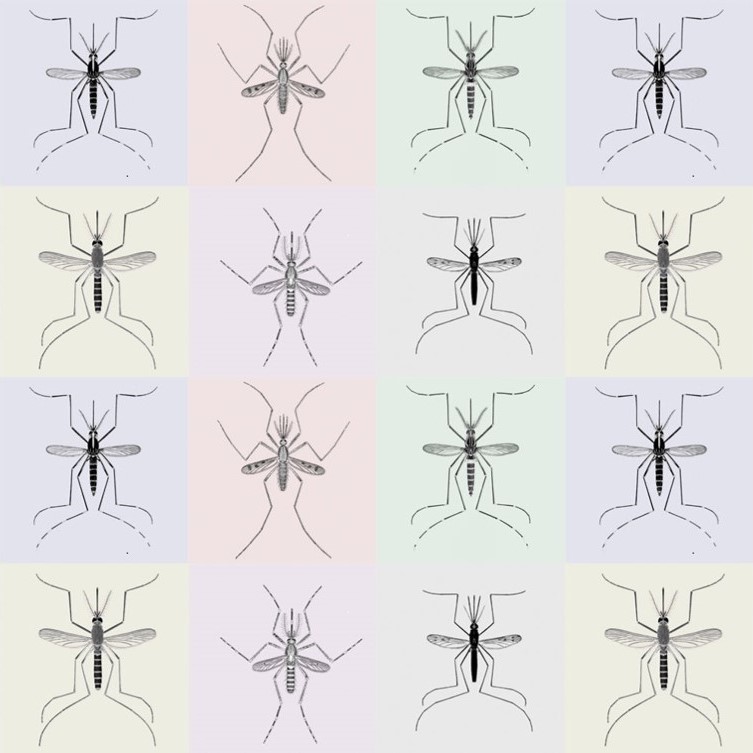}}
    \subfloat{\includegraphics[width=.5\columnwidth]{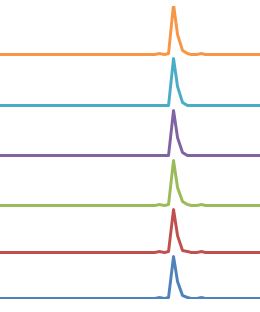}}
    \caption{Example of the first train case for the spectral problem InsectWingbeat.}
    \label{fig:InsectWingbeat}
\end{figure}

\subsection{Phoneme}

This dataset is a multivariate representation of a subset of the data used in the paper~\cite{hamooni14phoneme}.
Each series was extracted from the segmented audio collected from Google Translate. Audio files collected from Google translate are recorded at 22050 HZ. The speakers are male and female. After data collection, they segment waveforms of the words to generate phonemes using the Forced Aligner tool from the Penn Phonetics Laboratory. A Spectrogram of each instance was then created with a window size of 0.001 seconds and an overlap of 90\%.
Each instance in this multivariate dataset is arranged such that each dimension is a frequency band from the spectrogram. The data consists of 39 classes each with 170 instances.

\begin{figure}[!htb]
    \centering
    \subfloat{\includegraphics[trim=0cm -6cm 0cm 0cm, clip, width=.5\columnwidth]{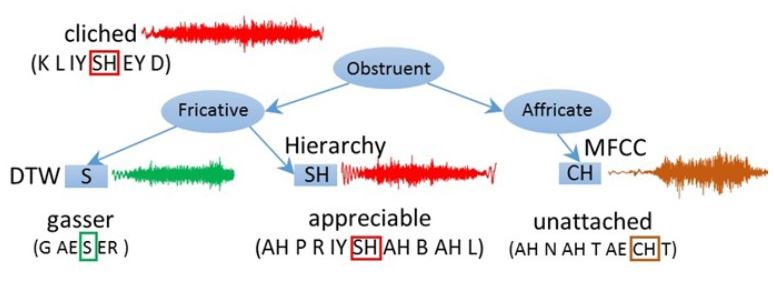}}
    \subfloat{\includegraphics[width=.5\columnwidth]{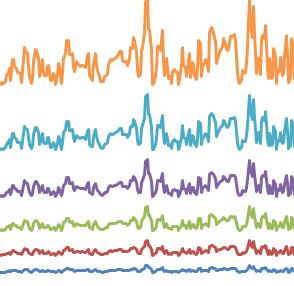}}
    \caption{Example of the first train case for the Audio problem Phoneme. }
    \label{fig:Phoneme}
\end{figure}

\subsection{SpokenArabicDigits}

This dataset is taken from the UCI repository. It is derived from sound.  8800 (10 digits x 10 repetitions x 88 speakers) samples were taken from 44 males and 44 females Arabic native speakers between the ages 18 and 40 to represent ten spoken Arabic digits. The 13 Mel Frequency Cepstral Coefficients (MFCCs) were computed with the following  conditions: Sampling rate 11025 Hz: 16 bits Hamming window; and filter pre-emphasized, $1-0.97Z^(-1)$~\cite{hammami09tree}.

\subsection{JapaneseVowels}

This dataset was taken from the  UCI Archive \footnote{https://archive.ics.uci.edu/ml/datasets/Japanese+Vowels}, originally reported in~\cite{kudo99multidimensional}.

\begin{figure}[!htb]
    \centering
    \subfloat{\includegraphics[width=.5\columnwidth]{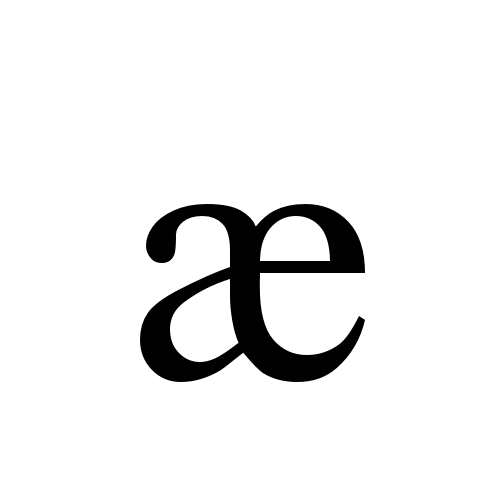}}
    \subfloat{\includegraphics[width=.5\columnwidth]{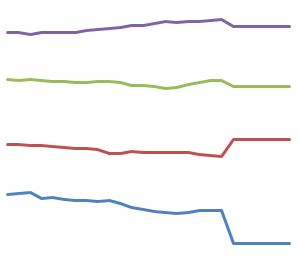}}
    \caption{Example of the first train case for the Audio problem JapaneseVowels. }
    \label{fig:JapaneseVowels}
\end{figure}
Nine Japanese-male speakers were recorded saying the vowels `a' and `e'. A `12-degree linear prediction analysis' is applied to the raw recordings to obtain time-series with 12 dimensions, a originally of lengths between 7 and 29. In this dataset, instances have been padded to the longest length; 29. The classification task is to predict the speaker. Therefore, each instance is a transformed utterance, 12*29 values with a single class label attached, 1\ldots9.

The given training set is comprised of 30 utterances for each speaker, however the test set has a varied distribution based on external factors of timing and experimental availability, between 24 and 88 instances per speaker.
\newpage
\section{Other Problems}

\subsection{EthanolConcentration}

\begin{figure}[!htb]
    \centering
    \subfloat{\includegraphics[trim=0cm -8cm 0cm 0cm, clip, width=.5\columnwidth]{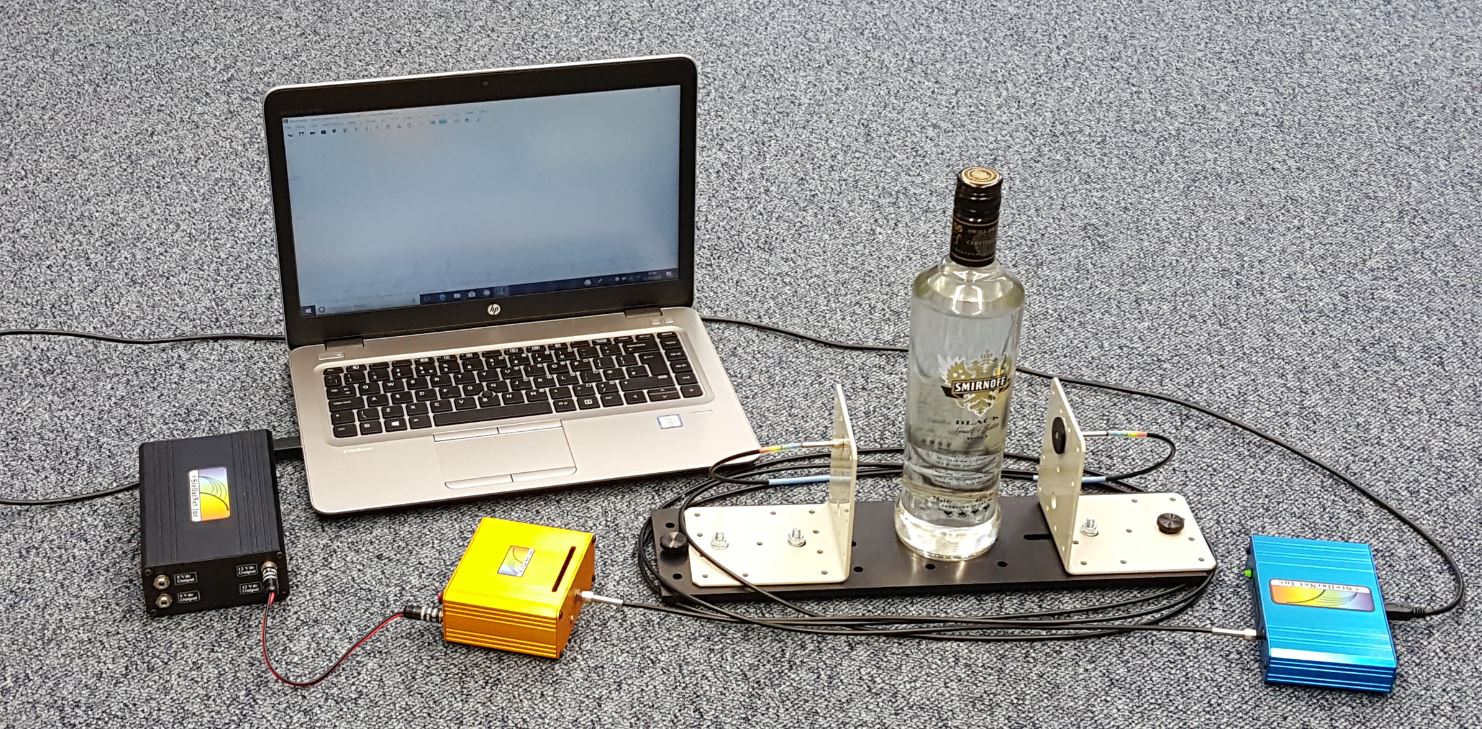}}
    \subfloat{\includegraphics[width=.5\columnwidth]{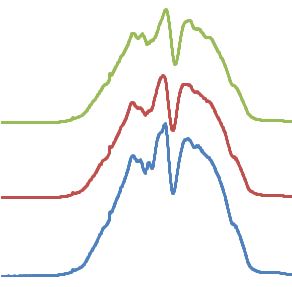}}
    \caption{Example of the first train case for the EEG problem EthanolConcentration. }
    \label{fig:EthanolConcentration}
\end{figure}

EthanolConcentration is a dataset of raw spectra of water-and-ethanol solutions in 44 distinct, real whisky bottles~\cite{large2018detecting}. The concentrations of ethanol are 35\%, 38\%, 40\%, and 45\%. The minimum legal alcohol limit for Scotch Whisky is 40\%, and many whiskies do maintain this alcohol concentration. Producers are required to ensure that the contents of their spirits contain alcohol concentrations that are tightly bound to what is reported on the labelling. The classification problem is to determine the alcohol concentration of a sample contained within an arbitrary bottle.

The data has been arranged such that each instance is made up of three repeat readings of the same bottle and batch of solution. Three solutions of each concentration (batches) were produced, and each bottle+batch combination measured three times. Each reading is comprised of the bottle being picked up, placed between the light source and spectroscope, and spectra saved. The spectra are recorded over the maximum wavelength range of the single StellarNet BLACKComet-SR spectrometer used (226nm to 1101.5nm with a sampling frequency of 0.5nm), over a one second integration time. Except for avoiding labelling, embossing, and seams on the bottle, no special attempts were made to obtain the cleanest reading for each individual bottle, nor to precisely replicate the exact path through the bottle for each repeat reading. This is to replicate potential future conditions of an operative performing mass-screening of a batch of suspect spirits.

Some bottles introduce more noise and structural defects to the spectra than others, based on their shape, colour, glass thickness and angle, and the ability to avoid the obstacles that may get in the way of a reading (labels, seams, etc). And so therefore the problem is to identify the alcohol concentration of the contents regardless of the properties of the containing bottle. 28 of the bottles are 'standard', that is, cylindrical with a roughly equal diameter, clear glass, with a clear path for the light to travel through. The remaining 16 bottles are either non-uniformly shaped, green glass, or light paths are difficult to find.

As well as the full dataset and an example 50/50 train test split, predefined folds in a 'leave one bottle out' format are given. All examples of a single bottle are reserved for the test set, meaning that the classifier cannot leverage the exact properties of the bottle of a new test sample already found in the training set.

For the problem's properties as a multivariate dataset, the dimensions are necessarily aligned in wavelength, and the relationship between them is moreso to allow for a noise cancelling or corrective affect, rather than each dimension describing strictly different information. Whether repeat readings and some form of multivariate method improves accuracy over classification on a single (univariate) reading is of interest. Interval methods are likely to provide benefits, as the wavelengths range from just into the Ultraviolet (UV) light, through the Visible (VIS) light, and into the Near Infrared (NIR). Different intervals carry different physical information.

\subsection{PEMS-SF}

This is a UCI dataset from the California Department of Transportation\footnote{www.pems.dot.ca.gov} reported in~\cite{cuturi11fast}.
It contains 15 months worth of daily data from the California Department of Transportation PEMS website. The data describes the occupancy rate, between 0 and 1, of different car lanes of San Francisco bay area freeways. The measurements cover the period from Jan. 1st 2008 to Mar. 30th 2009 and are sampled every 10 minutes. Each day in this database is a single time series of dimension 963 (the number of sensors which functioned consistently throughout the studied period) and length 6 x 24=144. Public holidays were removed from the dataset, as well
as two days with anomalies (March 8th 2009 and March 9th 2008) where all sensors were muted between 2:00 and 3:00 AM. This results in a database of 440 time series.

The task is to classify each observed day as the correct day of the week, from Monday to Sunday, e.g. label it with an integer in {1,2,3,4,5,6,7}.

\begin{figure}[!htb]
    \centering
    \subfloat{\includegraphics[trim=0cm -5cm 0cm 0cm, clip, width=.5\columnwidth]{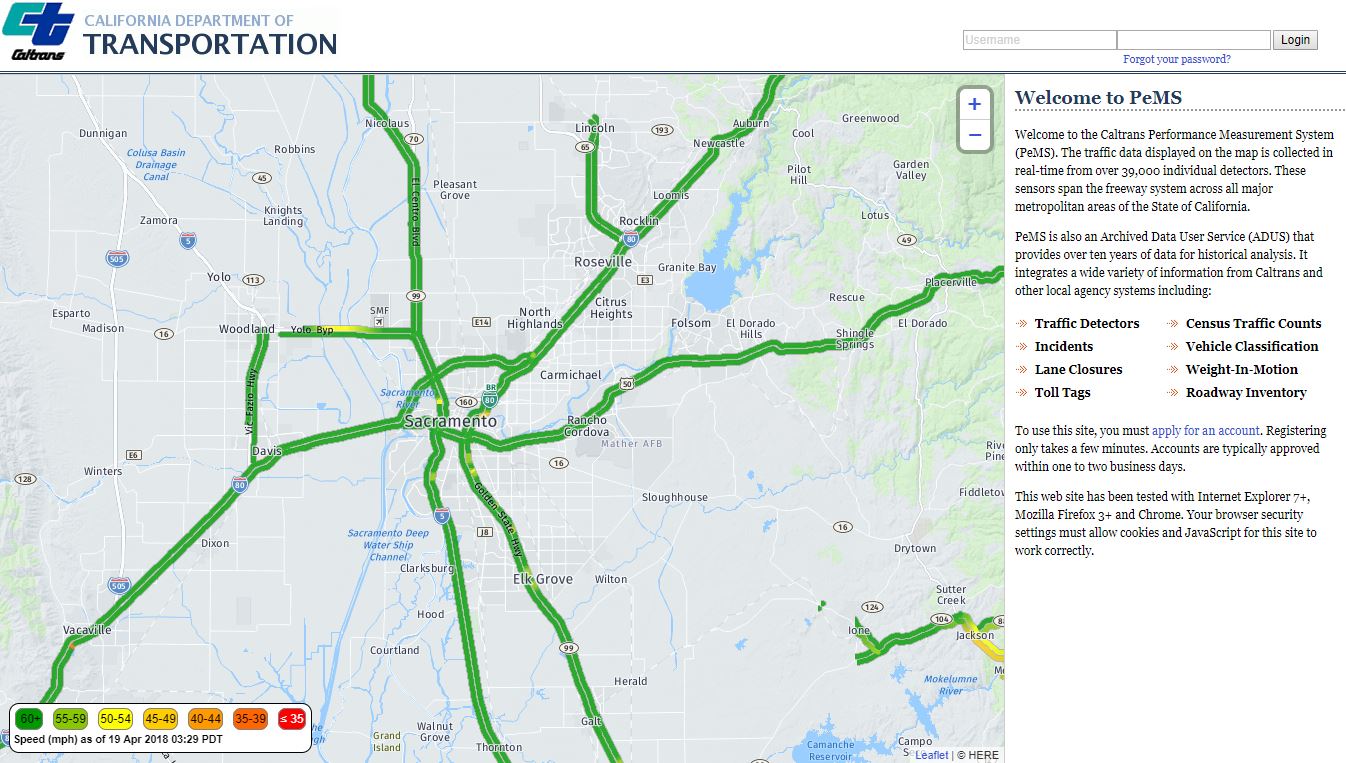}}
    \subfloat{\includegraphics[width=.5\columnwidth]{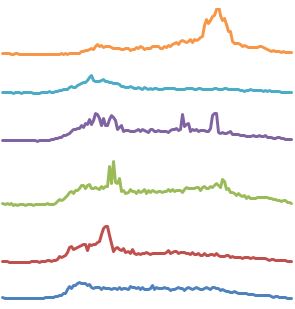}}
    \caption{Example of the first train case for the  problem PEMS-SF.}
    \label{fig:PEMS-SF}
\end{figure}

\subsection{LSST}

This dataset is from a 2018 Kaggle competition~\footnote{https://www.kaggle.com/c/PLAsTiCC-2018}.
The Photometric LSST Astronomical Time Series Classification
Challenge (PLAsTiCC) is an open data challenge to classify
simulated astronomical time-series data in preparation for
observations from the Large Synoptic Survey Telescope (LSST), which
will achieve first light in 2019 and commence its 10-year main
survey in 2022. LSST will revolutionize our understanding of the
changing sky, discovering and measuring millions of time-varying
objects.

PLAsTiCC is a large data challenge for which participants are asked
to classify astronomical time series data. These simulated time
series, or ‘light curves’ are measurements of an object’s
brightness as a function of time - by measuring the photon flux in
six different astronomical filters (commonly referred to as
passbands). These passbands include ultra-violet, optical and
infrared regions of the light spectrum. There are many different
types of astronomical objects (that are driven by different
physical processes) that we separate into astronomical classes.

The problem we have formulated represents a snap shot of the data available and is
created from the train set published in the aforementioned competition.

36 dimensions were chosen as it represents a value at which most
instances would not be truncated.

\begin{figure}[!htb]
    \centering
    \subfloat{\includegraphics[width=.5\columnwidth]{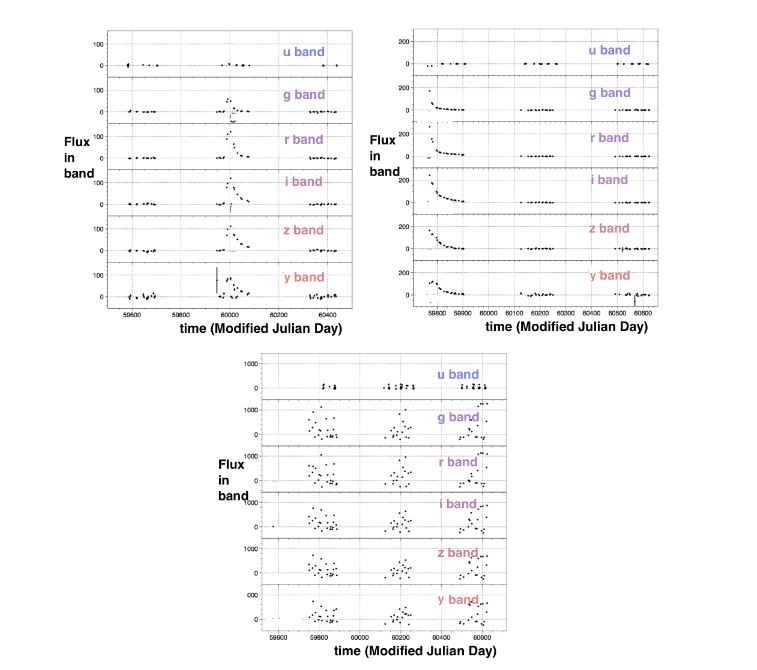}}
    \subfloat{\includegraphics[width=.5\columnwidth]{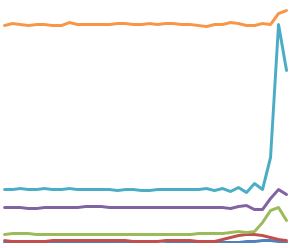}}
    \caption{Example of the first train case for the  problem LSST. }
    \label{fig:LSST}
\end{figure}
\newpage
\section{Benchmark Results}
\begin{figure}[!ht]
    \centering
    \includegraphics[trim = 3cm 11cm 2cm 1cm, clip, width=\linewidth]{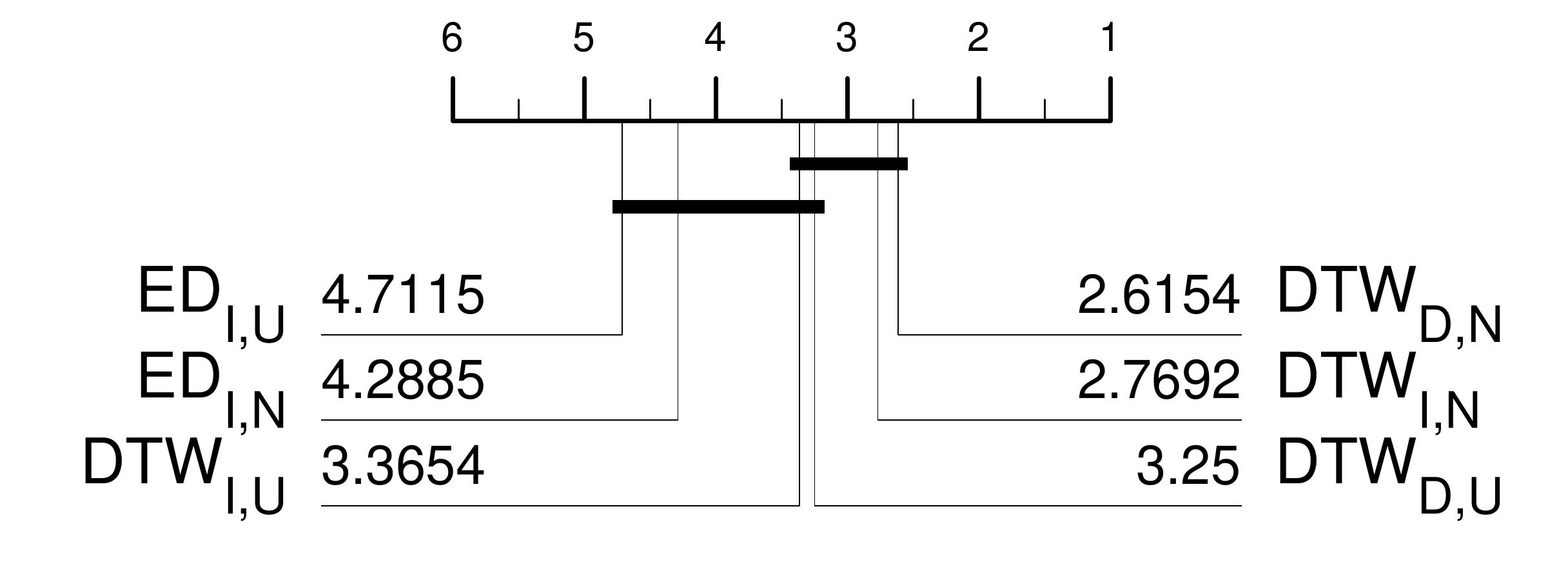}
        \caption{Critical Difference diagram of the benchmark classifiers across the datasets of the new archive. The subscripts `U' and `N' refer to the un-normalised and normalised versions of the classifiers, respectively.  The average accuracy ranking is given alongside the label of each classifier. Classifiers connected by a solid bar are not pairwise significantly different from each other.}
    \label{fig:benchmarksCDdia}
\end{figure}

Our initial benchmarking is with three standard classifiers for TSC: 1-Nearest Neighbour with distance functions: Euclidean (ED); dimension-independent dynamic time warping (DTW$_I$); and dimension-dependent dynamic time warping (DTW$_D$). A summary of the differences between these two multidimensional DTW variants can be found in~\cite{shokoohi15non}. We present results using the raw data, and after normalising each dimension independently. Accuracies are presented in Table~\ref{tab:benchmarks}, and these are summarised in a critical difference diagram in Figure~\ref{fig:benchmarksCDdia}. At the time of release we do not have full results for three datasets: EigenWorms; InsectWingbeat and FaceDetection. We will add these when complete. Full results will also be on the website~\cite{tscWeb}.

We can see that a wide range of performances are achieved by the benchmarks. Five of the datasets can be classified with one or more of the benchmark classifiers to at least 99\% accuracy, and five cannot do better than 50\%. Trivial or impossible problems may be removed from the archive in future iterations, depending on a wider scale performance evaluation.  

\begin{table}[]
    \centering
    \resizebox{\linewidth}{!}{
        \begin{tabular}{l|ccc|ccc}
            \hline
            & \multicolumn{3}{c|}{Un-normalised} & \multicolumn{3}{c}{Normalised} \\
	                                    & ED$_I$	    & DTW$_I$	    & DTW$_D$       & ED$_I$	    & DTW$_I$	    & DTW$_D$       \\
	        \hline
            ArticularyWordRecognition	& 0.97	        & 0.98	        & \textbf{0.987} & 0.97	        & 0.98	        & \textbf{0.987}\\
            AtrialFibrillation	        & \textbf{0.267}	        & \textbf{0.267}	        & 0.2	        & \textbf{0.267}         & \textbf{0.267}	        & 0.22	        \\
            BasicMotions	            & 0.675	        & \textbf{1}	& 0.975         & 0.676	        & \textbf{1} 	        & 0.975	        \\
            CharacterTrajectories	    & 0.964	        & 0.969	        & \textbf{0.99} & 0.964	        & 0.969	        & 0.989	        \\
            Cricket	                    & 0.944	        & 0.986	        & \textbf{1}    & 0.944	        & 0.986	        & \textbf{1}	\\
            DuckDuckGeese	            & 0.275	        & 0.55	        & \textbf{0.6}  & 0.275	        & 0.55	        & \textbf{0.6}	        \\
            EigenWorms                  & 0.550	        & 0.603	        & \textbf{0.618}& 0.549         &               & \textbf{0.618}         \\
            Epilepsy	                & 0.667	        & \textbf{0.978}& 0.964         & 0.666	        & \textbf{0.978}& 0.964	        \\
            ERing	                    & \textbf{0.133}	        & \textbf{0.133}	        & \textbf{0.133}         & \textbf{0.133}	        & \textbf{0.133}& \textbf{0.133}	        \\
            EthanolConcentration	    & 0.293	        & 0.304	        & \textbf{0.323}& 0.293	        & 0.304	        & \textbf{0.323}         \\
            FaceDetection               & 0.519	        & 0.513	        & \textbf{0.529}& 0.519              &               & \textbf{0.529}              \\
            FingerMovements	            & \textbf{0.55}	        & 0.52	        & 0.53          & \textbf{0.55}	        & 0.52 & 0.53	        \\
            HandMovementDirection	    & 0.279	        & \textbf{0.306}	        & 0.231         & 0.278	        & \textbf{0.306}	        & 0.231\\
            Handwriting	                & 0.371	        & 0.509	        & \textbf{0.607}& 0.2	        & 0.316	        & 0.286	        \\
            Heartbeat	                & 0.62	        & 0.659	        & \textbf{0.717}& 0.619	        & 0.658	        & \textbf{0.717}	        \\
            InsectWingbeat              & \textbf{0.128}&               & 0.115         & \textbf{0.128}         &               &               \\
            JapaneseVowels	            & 0.924	        & \textbf{0.959}	        & 0.949         & 0.924	        & \textbf{0.959}& 0.949	         \\
            Libras	                    & 0.833	        & \textbf{0.894}& 0.872         & 0.833         & \textbf{0.894}	        & 0.87	        \\
            LSST	                    & 0.456	        & \textbf{0.575}& 0.551         & 0.456	        & \textbf{0.575}	        & 0.551	        \\
            MotorImagery                & \textbf{0.51}& 0.39	        & 0.5         & \textbf{0.51}          &               & 0.5           \\
            NATOPS	                    & 0.85	        & 0.85	        & \textbf{0.883}& 0.85	        & 0.85	        & \textbf{0.883}	        \\
            PEMS-SF	                    & 0.705	        & \textbf{0.734}	        & 0.711         & 0.705	        & \textbf{0.734}& 0.711	        \\
            PenDigits	                & 0.973	        & 0.939	        & \textbf{0.977}         & 0.973	        & 0.939	        & \textbf{0.977}\\
            Phoneme	                    & 0.104	        & \textbf{0.151}	        & \textbf{0.151}         & 0.104	        & \textbf{0.151}& \textbf{0.151}\\
            RacketSports	            & \textbf{0.868}& 0.842	        & 0.803         & \textbf{0.868}	        & 0.842	        & 0.803        \\
            SelfRegulationSCP1	        & 0.771	        & 0.765	        & \textbf{0.775}         & 0.771	        & 0.765	        & \textbf{0.775}\\
            SelfRegulationSCP2	        & 0.483	        & 0.533	        & \textbf{0.539}& 0.483	        & 0.533	        & \textbf{0.539}	        \\
            SpokenArabicDigits	        & \textbf{0.967}	        & 0.96	        & 0.963         & \textbf{0.967}	        & 0.959	        & 0.963\\
            StandWalkJump	            & 0.2           & \textbf{0.333}& 0.2           & 0.2	        & \textbf{0.333}	        & 0.2	        \\
            UWaveGestureLibrary	        & 0.881	        & 0.869	        & \textbf{0.903}         & 0.881	        & 0.868        & \textbf{0.903}\\
            \hline
        \end{tabular}
    } 
    \caption{Benchmark classification results (in terms of accuracy) for the original and normalised versions of each dataset in the new archive. The (potentially tied) best accuracy achieved for a dataset is in bold.}
    \label{tab:benchmarks}
\end{table}

These results are our first attempt at benchmarking. We will expand these results over the ensuing months. We will also conduct resample and/or cross validation experiments.

\newpage
\section{Conclusions}

This is our first attempt at a multivariate archive, and it should be considered a work in progress. We hope to release an expanded version in 2019. We would very much welcome any donations of data. If you have evaluated your classifier on this data, your results are reproducable and your work has been peer reviewed, get in touch and we will put your results and algorithm details on the website. If you find any errors in the data or the descriptions, please inform us.


\begin{thebibliography}{10}

\bibitem{dau18archive}
H.~Dau, A.~Bagnall, K.~Kamgar, M.~Yeh, Y.~Zhu, S.~Gharghabi, and
  C.~Ratanamahatana, ``The ucr time series archive,'' {\em ArXiv e-prints},
  vol.~arXiv:1810.07758, 2018.

\bibitem{bagnall17bakeoff}
A.~Bagnall, J.~Lines, A.~Bostrom, J.~Large, and E.~Keogh, ``The great time
  series classification bake off: a review and experimental evaluation of
  recent algorithmic advances,'' {\em Data Mining and Knowledge Discovery},
  vol.~31, no.~3, pp.~606--660, 2017.

\bibitem{ko2005online}
M.~H. Ko, G.~West, S.~Venkatesh, and M.~Kumar, ``Online context recognition in
  multisensor systems using dynamic time warping,'' in {\em Intelligent
  Sensors, Sensor Networks and Information Processing Conference, 2005.
  Proceedings of the 2005 International Conference on}, pp.~283--288, IEEE,
  2005.

\bibitem{shokoohi17generalizing}
M.~Shokoohi-Yekta, B.~Hu, H.~Jin, J.~Wang, and E.~Keogh, ``Generalizing {DTW}
  to the multi-dimensional case requires an adaptive approach,'' {\em Data
  Mining and Knowledge Discovery}, vol.~31, no.~1, pp.~1--31, 2017.

\bibitem{villar2016generalized}
J.~R. Villar, P.~Vergara, M.~Men{\'e}ndez, E.~de~la Cal, V.~M. Gonz{\'a}lez,
  and J.~Sedano, ``Generalized models for the classification of abnormal
  movements in daily life and its applicability to epilepsy convulsion
  recognition,'' {\em International journal of neural systems}, vol.~26,
  no.~06, p.~1650037, 2016.

\bibitem{wilhelm2015ering}
M.~Wilhelm, D.~Krakowczyk, F.~Trollmann, and S.~Albayrak, ``ering: multiple
  finger gesture recognition with one ring using an electric field,'' in {\em
  Proceedings of the 2nd international Workshop on Sensor-based Activity
  Recognition and Interaction}, p.~7, ACM, 2015.

\bibitem{dias16algoritmos}
D.~B. Dias and S.~M. Peres, ``Algoritmos bio-inspirados aplicados ao
  reconhecimento de padroes da libras: enfoque no par{\^a}metro movimento,''
  {\em 16 ˆAo Simp{\'o}sio Internacional de Inicia{\c{c}}ao Cient{\i}fica da
  Universidade de Sao Paulo}, 2016.

\bibitem{ghouaiel2017continuous}
N.~Ghouaiel, P.-F. Marteau, and M.~Dupont, ``Continuous pattern detection and
  recognition in stream-a benchmark for online gesture recognition,'' {\em
  International Journal of Applied Pattern Recognition}, vol.~4, no.~2, 2017.

\bibitem{liu2009uwave}
J.~Liu, L.~Zhong, J.~Wickramasuriya, and V.~Vasudevan, ``uwave:
  Accelerometer-based personalized gesture recognition and its applications,''
  {\em Pervasive and Mobile Computing}, vol.~5, no.~6, pp.~657--675, 2009.

\bibitem{wang2013word}
J.~Wang, A.~Balasubramanian, L.~M. de~La~Vega, J.~R. Green, A.~Samal, and
  B.~Prabhakaran, ``Word recognition from continuous articulatory movement
  time-series data using symbolic representations,'' in {\em Proceedings of the
  Fourth Workshop on Speech and Language Processing for Assistive
  Technologies}, pp.~119--127, 2013.

\bibitem{williams2008modelling}
B.~Williams, M.~Toussaint, and A.~J. Storkey, ``Modelling motion primitives and
  their timing in biologically executed movements,'' in {\em Advances in neural
  information processing systems}, pp.~1609--1616, 2008.

\bibitem{brown2013dictionary}
A.~E. Brown, E.~I. Yemini, L.~J. Grundy, T.~Jucikas, and W.~R. Schafer, ``A
  dictionary of behavioral motifs reveals clusters of genes affecting
  caenorhabditis elegans locomotion,'' {\em Proceedings of the National Academy
  of Sciences}, vol.~110, no.~2, pp.~791--796, 2013.

\bibitem{yemini2013database}
E.~Yemini, T.~Jucikas, L.~J. Grundy, A.~E. Brown, and W.~R. Schafer, ``A
  database of caenorhabditis elegans behavioral phenotypes,'' {\em Nature
  methods}, vol.~10, no.~9, p.~877, 2013.

\bibitem{alimouglu01combining}
F.~Alimo{\u{g}}lu and E.~Alpaydin, ``Combining multiple representations for
  pen-based handwritten digit recognition,'' {\em Turkish Journal of Electrical
  Engineering \& Computer Sciences}, vol.~9, no.~1, pp.~1--12, 2001.

\bibitem{moody2004spontaneous}
G.~Moody, ``Spontaneous termination of atrial fibrillation: a challenge from
  physionet and computers in cardiology 2004,'' in {\em Computers in
  Cardiology, 2004}, pp.~101--104, IEEE, 2004.

\bibitem{Behravan2015Rate}
V.~Behravan, N.~E. Glover, R.~Farry, P.~Y. Chiang, and M.~Shoaib,
  ``Rate-adaptive compressed-sensing and sparsity variance of biomedical
  signals,'' in {\em 2015 IEEE 12th International Conference on Wearable and
  Implantable Body Sensor Networks (BSN)}, pp.~1--6, June 2015.

\bibitem{blankertz2002classifying}
B.~Blankertz, G.~Curio, and K.-R. M{\"u}ller, ``Classifying single trial eeg:
  Towards brain computer interfacing,'' in {\em Advances in neural information
  processing systems}, pp.~157--164, 2002.

\bibitem{lal05methods}
T.~Lal, T.~Hinterberger, G.~Widman, M.~Schr{\"o}der, N.~J. Hill, W.~Rosenstiel,
  C.~E. Elger, N.~Birbaumer, and B.~Sch{\"o}lkopf, ``Methods towards invasive
  human brain computer interfaces,'' in {\em Advances in neural information
  processing systems}, pp.~737--744, 2005.

\bibitem{birbaumer99spelling}
N.~Birbaumer, N.~Ghanayim, T.~Hinterberger, I.~Iversen, B.~Kotchoubey,
  A.~K{\"u}bler, J.~Perelmouter, E.~Taub, and H.~Flor, ``A spelling device for
  the paralysed,'' {\em Nature}, vol.~398, no.~6725, p.~297, 1999.

\bibitem{chen2014flying}
Y.~Chen, A.~Why, G.~Batista, A.~Mafra-Neto, and E.~Keogh, ``Flying insect
  classification with inexpensive sensors,'' {\em Journal of insect behavior},
  vol.~27, no.~5, pp.~657--677, 2014.

\bibitem{hamooni14phoneme}
H.~Hamooni and A.~Mueen, ``Dual-domain hierarchical classification of phonetic
  time series,'' in {\em Proc. {IEEE} International Conference on Data Mining},
  2014.

\bibitem{hammami09tree}
N.~Hammami and M.~Sellam, ``Tree distribution classifier for automatic spoken
  arabic digit recognition,'' in {\em Internet Technology and Secured
  Transactions, 2009. ICITST 2009. International Conference for}, pp.~1--4,
  IEEE, 2009.

\bibitem{kudo99multidimensional}
M.~Kudo, J.~Toyama, and M.~Shimbo, ``Multidimensional curve classification
  using passing-through regions,'' {\em Pattern Recognition Letters}, vol.~20,
  no.~11, pp.~1103--1111, 1999.

\bibitem{large2018detecting}
J.~Large, E.~K. Kemsley, N.~Wellner, I.~Goodall, and A.~Bagnall, ``Detecting
  forged alcohol non-invasively through vibrational spectroscopy and machine
  learning,'' in {\em Pacific-Asia Conference on Knowledge Discovery and Data
  Mining}, pp.~298--309, Springer, 2018.

\bibitem{cuturi11fast}
M.~Cuturi, ``Fast global alignment kernels,'' in {\em Proceedings of the 28th
  international conference on machine learning (ICML-11)}, pp.~929--936, 2011.

\bibitem{shokoohi15non}
M.~Shokoohi-Yekta, J.~Wang, and E.~Keogh, ``On the non-trivial generalization
  of dynamic time warping to the multi-dimensional case,'' in {\em Proceedings
  of the 2015 SIAM International Conference on Data Mining}, pp.~289--297,
  SIAM, 2015.

\bibitem{tscWeb}
A.~Bagnall, J.~Lines, and E.~Keogh, ``The {UEA UCR} time series classification
  archive.'' {\tt http://timeseriesclassification.com}, 2018.

\end{thebibliography}
\end{document}